\pdfoutput=1

\documentclass[11pt]{article}

\usepackage[final]{coling}

\usepackage{times}
\usepackage{latexsym}

\usepackage[T1]{fontenc}

\usepackage[utf8]{inputenc}

\usepackage{microtype}

\usepackage{inconsolata}

\usepackage{graphicx}
\usepackage{array}
\usepackage{amsmath}
\usepackage{bm}
\usepackage{algorithm}
\usepackage{algpseudocode}
\usepackage{amsmath}
\usepackage{multirow}
\usepackage{indentfirst}
\usepackage{subfigure}
\usepackage{epsfig}
\usepackage{epstopdf}
\usepackage{graphicx}
\usepackage{url}
\usepackage{xspace}
\usepackage{booktabs}
\usepackage{subeqnarray}
\usepackage{subcaption}
\usepackage{color}

\newcommand{\paratitle}[1]{\vspace{1.5ex}\noindent\textbf{#1}}
\newcommand{\ie}{\emph{i.e.,}\xspace}

\newcommand{\eg}{\emph{e.g.,}\xspace}

\newcommand{\ignore}[1]{}

\title{Investigating the Factual Knowledge Boundary of Large Language Models with Retrieval Augmentation}

\author{\textbf{Ruiyang Ren\textsuperscript{1,3}\thanks{~~Equal contributions.}\thanks{~~The work was done during the internship at Baidu.}\quad 
Yuhao Wang\textsuperscript{1,3}\footnotemark[1] \quad
Yingqi Qu\textsuperscript{2} \quad
Wayne Xin Zhao\textsuperscript{1,3}\thanks{\llap{}\:\:\:Corresponding authors. } \quad 
Jing Liu\textsuperscript{2}\footnotemark[3]}  \\
\textbf{Hao Tian\textsuperscript{2} \quad
Hua Wu\textsuperscript{2} \quad
Ji-Rong Wen\textsuperscript{1,3} \quad  Haifeng Wang\textsuperscript{2}  
}\\
	\textsuperscript{1}Gaoling School of Artificial Intelligence, Renmin University of China  \\
	\textsuperscript{2}Baidu Inc. \\
	\textsuperscript{3}Beijing Key Laboratory of Big Data Management and Analysis Methods\\
	\{reyon.ren, yh.wang, jrwen\}@ruc.edu.cn, batmanfly@gmail.com\\
	\{quyingqi, liujing46, tianhao, wu\_hua, wanghaifeng\}@baidu.com
}

\begin{document}
\maketitle
\begin{abstract}
Large language models (LLMs) have shown impressive prowess in solving a wide range of tasks with world knowledge. However, it remains unclear how well LLMs are able to perceive their factual knowledge boundaries, particularly under retrieval augmentation settings.
In this study, we present the first analysis on the factual knowledge boundaries of LLMs and how retrieval augmentation affects LLMs on open-domain question answering~(QA), with a bunch of important findings.
Specifically, we focus on three research questions and analyze them by examining QA, priori judgement and posteriori judgement capabilities of LLMs. 
We show evidence that LLMs possess unwavering confidence in their knowledge and cannot handle the conflict between internal and external knowledge well. Furthermore, retrieval augmentation proves to be an effective approach in enhancing LLMs' awareness of knowledge boundaries. We further conduct thorough experiments to examine how different factors affect LLMs and propose a simple method to dynamically utilize supporting documents with our judgement strategy.
Additionally, we find that the relevance between the supporting documents and the questions significantly impacts LLMs' QA and judgemental capabilities. The code to reproduce this work is available at \url{https://github.com/RUCAIBox/LLM-Knowledge-Boundary}. 
\end{abstract}

\section{Introduction}

Knowledge-intensive tasks, defined by their requirement for extensive knowledge, represent a significant area of interest within the field of natural language processing~\citep{petroni2020kilt}. 
A prime example of such tasks is open-domain question answering (QA)~\citep{chen17openqa},
which necessitates the assistance of information retrieval systems~\citep{zhao2023survey} to obtain relevant information.
Subsequently, a reading comprehension model is employed to identify and retrieve pertinent information, ultimately yielding the final response.

Recently, large language models~(LLMs) have showcased remarkable abilities in solving various tasks, which are capable of encoding extensive volumes of world knowledge within their parameters~\citep{brown2020language, Ouyang-arxiv-2022-Training, zhao2023survey}. 
Despite their exceptional capabilities, LLMs may exhibit limited flexibility in knowledge-intensive tasks, necessitating the incorporation of retrieval augmentation strategies. Several pioneering efforts have applied LLMs to open-domain QA tasks~\citep{qin2023chatgpt, kamalloo2023evaluating, wang2023evaluating, sun2023beamsearchqa, xu2024unsupervised}.
Typically, they mainly focus on evaluating the QA performance of LLMs, discussing improved evaluation methods or leveraging LLMs to enhance QA models. The existing effort also detects the uncertainty of LLMs with an automated method~\citep{YinSGWQH23}.
Furthermore,
the issue of hallucination poses challenges to LLMs' reliable deployment, while retrieval augmentation is considered an effective method to mitigate hallucinations~\citep{li2024dawn, shuster2021retrieval, wang2023hallucination}.

Despite the contribution of existing studies, there is still a lack of a deep understanding of LLMs' capabilities in perceiving their factual knowledge boundaries, particularly when external resources can be used.
For instance, it remains uncertain whether LLMs are able to assess their own ability to answer questions, evaluate whether the provided reference is sufficient to address the questions, and assess the accuracy of their own answers.
Our primary focus is the factual knowledge boundary of LLMs, and study the impact of retrieval augmentation on the generation of LLMs.

To this end, we undertake a thorough analysis of the influence of retrieval augmentation on the generation quality of LLMs, with a specific focus on QA performance and LLMs' perception of their factual knowledge boundaries. 
In order to fully explore the knowledge boundaries of LLMs, we consider a wide range of LLMs, including closed source LLMs and publicly available LLMs.
To measure the capacity of knowledge boundary perception, we consider two alternative approaches. The first one is  \emph{priori judgement}, in which LLMs assess the feasibility of answering a given question before the response generation process. The second one is \emph{posteriori judgement}, where LLMs evaluate the correctness of their answer to questions after the response generation process. The two approaches can evaluate the knowledge boundary perception of LLMs from different perspectives.
For retrieval augmentation, we adopt multiple retrieval models to provide supporting documents for LLMs regarding the given questions.

To conduct a comprehensive investigation, our work aims to answer three research questions progressively: 
{(i) To what extent can LLMs perceive their factual knowledge boundaries?}
{(ii) What effect does retrieval augmentation have on LLMs?}
{(iii) How do supporting documents with different characteristics affect LLMs? }

We design comprehensive experiments to thoroughly explore these research questions individually. Furthermore, building upon priori judgement, we attempt a simple method for dynamically introducing retrieval augmentation to LLMs with a performance gain.
Based on the empirical analysis, we have derived the following important findings corresponding to the three research questions:

\begin{itemize}
\item LLMs’ perception of the factual knowledge boundary is inaccurate and they often display a tendency towards being overconfident. LLMs are also not able to handle internal-external knowledge conflicts well.

\item LLMs cannot sufficiently utilize their internal knowledge, while retrieval augmentation can provide a beneficial knowledge supplement for LLMs, especially for smaller LLMs. 
Retrieval augmentation can also enhance LLMs' perception on their factual knowledge boundaries. 
Furthermore, the performance of retrieval-augmented LLMs is affected by model-specific factors, question types, and the characteristics of the retrieval documents.

\item LLMs exhibit improved performance and confidence when referring high-quality supporting documents and tend to rely on the provided information to produce the responses. The reliance extent and LLMs' confidence are contingent upon the relevance between the supporting documents and the question.
\end{itemize}

\section{Background and Methodology}
In this section, we provide an overview of the background and fundamental methodologies that are essential for this study.

\subsection{Task Formulation}
We conduct experiments on open-domain question answering~(QA), which can be described as follows. Given a question $q$ in natural language and a large document collection $\mathcal{D}=\{ d_i \}_{i=1}^m$ such as Wikipedia, the model is required to provide an answer $a$ to the question $q$ with the help of the provided corpus $\mathcal{D}$. 
With large language models~(LLMs), the open-domain QA task can be directly solved in an end-to-end manner in a closed-book setting~\citep{qin2023chatgpt}. Given a question $q$, the answer $a$ can be generated by the LLM with a prompt $p$ following a specific output format:
\begin{equation}
    a = f_\text{LLM}(p, q).
    \label{eq:llm}
\end{equation}
When enhancing the LLM with retrieval, a typical strategy is designing prompt $p$ to instruct the LLM to provide an answer $a$ to question $q$ using the supporting documents $\mathcal{L}$ retrieved by the retriever:
\begin{equation}
    a = f_\text{LLM}(p, q, \mathcal{L}).
    \label{eq:retrieval}
\end{equation}
Equation~\ref{eq:llm} and \ref{eq:retrieval} present two approaches for LLMs to solve QA tasks, we also adopt these approaches to evaluate LLMs' factual knowledge boundaries.

\subsection{Instructing LLMs with Natural Language Prompts}
\label{sec:prompting}
\begin{figure*}
    \centering
    \includegraphics[width=0.77\textwidth]{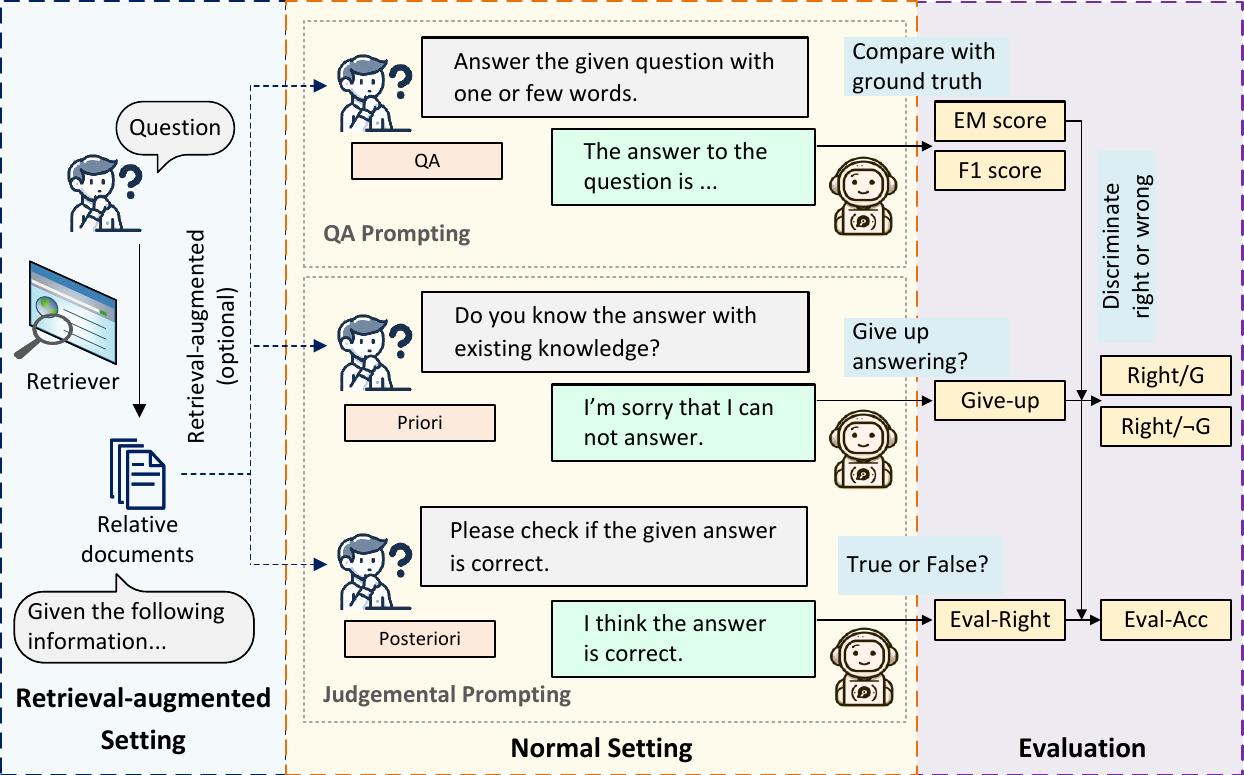}
    \caption{The illustration of different settings to instruct LLMs, the evaluation metrics are also displayed.}
    \label{fig:settings}
\end{figure*}

We consider two instruction settings, namely QA prompting and judgemental prompting. 
Figure~\ref{fig:settings} provides an overall illustration of our prompting strategies and detailed instructions can be found in Appendix~\ref{appendix:prompts}.

\subsubsection{QA Prompting}

To evaluate LLMs' ability to utilize knowledge, we focus LLMs' QA ability that provides answers to the given questions, with two-fold aims. Firstly, we manage to ensure that LLMs can accurately answer and refer to both internal and external knowledge. Secondly, we encourage concise responses to match the short-answer format for assessment with normally used metrics.
To this end, we propose two distinct instructional approaches.
(1) \textit{Normal setting:} In this approach, LLMs are instructed to respond to questions based solely on their internal knowledge, as formulated in Equation~(\ref{eq:llm}). 
(2) \textit{Retrieval-augmented setting:} This approach requires LLMs to answer the given questions using a combination of their internal knowledge and information from supporting documents retrieved, as outlined in Equation~(\ref{eq:retrieval}). In this setting, supporting documents are optional, and ideally, LLMs should determine whether to refer to the supporting documents based on their reliability.

\subsubsection{Judgemental Prompting} 
To investigate whether LLMs are capable of perceiving their own factual knowledge boundary, we propose judgemental prompting to evaluate the judging abilities of LLMs. 

Similar to QA prompting, the concepts of the {\textit{normal setting}} and the {\textit{retrieval-augmented setting}} are also applicable for judgemental prompting, where LLMs utilizing their own knowledge or consulting supporting documents from retrievers to carry out the judgement process.
Furthermore, we construct instructions from different judgement perspectives.
(1) \textit{Priori judgement:} LLMs are tasked with determining their capability to provide an answer to the question, employing either the normal setting or the retrieval-augmented setting. 
Priori judgment is carried out before the answering of LLMs, through this predictive approach, we are able to evaluate the confidence levels and assessment accuracies of LLMs in their mastery of knowledge.
(2) \textit{Posteriori judgement:} LLMs are tasked with evaluating the correctness of the answer to the question provided by itself, employing either a normal setting or a retrieval-augmented setting. 
Posteriori judgement aims to enable LLMs to judge their responses. Through this reflective approach, we can evaluate the confidence levels and assessment accuracies of LLMs in the content they generate.

\subsection{Experimental Settings}
In this part, we set up our experiments on open-domain QA, including evaluation metrics and retrieval sources. Due to the space limitaion, more settings can be found in Appendix~\ref{appendix:more-settings}, including datasets, evaluation models, implemental details and instruction designs.

\subsubsection{Evaluation Metrics}
\label{sec:metrics}
Following previous works~\citep{chen17openqa, Izacard2021LeveragingPR, sun2023beamsearchqa}, we use the \textit{exact match~(EM)} score and \textit{F1} score to evaluate the QA performance of LLMs. 
\textit{EM} calculates the percentage of questions in which the answer predicted by LLMs precisely matches the correct answer to the question. 
\textit{F1} is used to measure the overlap between the predicted answer and the correct answer, combines precision and recall into a single metric by taking their harmonic mean. 

We also propose several metrics for evaluating LLMs’ judgement abilities.
\textit{Give-up} denotes the percentage of questions that LLMs give up answering, which reflects the LLMs' judgement on whether they possess the relevant knowledge.
\textit{Right/G} represents the probability that LLMs give up answering but can actually answer correctly. \textit{Right/$\neg$G} represents the probability that LLMs do not give up answering and can answer correctly. 
\textit{Eval-Right} refers to the proportion of questions where LLMs assess their answers as correct. 
\textit{Eval-Acc} represents the percentage of questions for which the assessment of the answer by LLMs aligns with the fact.
Among them, Give-up, Right/G and Right/$\neg$G are metrics for \textit{priori judgement}, Eval-Right and Eval-Acc are metrics for \textit{posteriori judgement}. All metrics are illustrated in Figure~\ref{fig:settings}.

\subsubsection{Retrieval Sources}
\label{sec:retrieval-sources}
We consider multiple retrieval sources to acquire supporting documents, including dense retrieval~\citep{gao2021unsupervised, pair, Zhuang2022ImplicitFF, Zhou2022SimANS}, sparse retrieval~\citep{robertson2009probabilistic} and ChatGPT.

For the dense retriever, we utilize RocketQAv2 \citep{rocketqav2} to find semantically relevant documents for questions. To achieve this, we train the model on each dataset with the constructed in-domain training data under the settings of RocketQAv2 and leverage Faiss~\citep{faiss} to obtain relevant documents for each question from the candidate corpus.
For the sparse retriever, we use BM25~\citep{Yang2017AnseriniET} to find lexical relevant documents for questions. 
Similar to previous works~\citep{yu2022generate, ren2023tome}, we regard the generative language model as a ``retriever'' that ``retrieves'' knowledge from its memory, where ChatGPT is instructed to produce relevant documents in response to a given question.
The retrieval results can be found in Appendix~\ref{appendix:retrieval_results}.

\section{Experimental Analysis and Findings}
\label{sec:analysis}
In this section, we focus on addressing the three research questions and 
tackle them by investigating the \textit{judgement ability} and the \textit{QA ability} of LLMs with the proposed strategies in Section~\ref{sec:prompting}.

\subsection{To What Extent Can LLMs Perceive Their Factual Knowledge Boundaries?}
We deconstruct the question and investigate the following points: (a) Before answering: how do LLMs decide to abstain from a question? (b) When answering: can LLMs accurately address a given question? (c) After answering: how do LLMs assess the correctness of their answers?

\subsubsection{Settings}
In this part of the experiments, we employ the priori judgement with the normal setting to instruct LLMs on the decision of whether to give up answering questions based on their inherent knowledge, and we use the QA prompting with the normal setting to instruct LLMs to provide answers. Moreover, we employ posteriori judgement with the normal setting to instruct LLMs in evaluating the correctness of their answers.

\begin{table}[t]
\centering
\scriptsize
\renewcommand\tabcolsep{2.3pt}
\begin{tabular}{lccccccc}
    \toprule
    \multirow{2.5}{*}{\textbf{}} & \multirow{2.5}{*}{\textbf{LLM}} & \multicolumn{2}{c}{{QA}} & \multicolumn{2}{c}{{Priori Judgement}} & \multicolumn{2}{c}{{Posteriori Judgement}}\\
    \cmidrule(lr){3-4}
    \cmidrule(lr){5-6}
    \cmidrule(lr){7-8}
    &&\textbf{EM} & \textbf{F1} & \textbf{Give-up}  & \textbf{Right/$\neg$G} & \textbf{Eval-Right} & \textbf{Eval-Acc}\\
    \midrule
    \multirow{5}{*}{\rotatebox[origin=c]{90}{\textbf{NQ}}} & Davinci003  &{27.20} & {36.20} & 29.20\% &  32.77\% & 72.80\% & 45.01\%\\
    & ChatGPT & {33.40} & {45.32} & 57.40\% & 42.72\% & 84.40\% & 43.40\%\\
    & GPT-4 & {34.60} & {48.72} & 15.20\% & 39.15\% & 90.20\% & 38.87\%\\
    & LLaMA2 & {16.60} & {24.26} & 6.60\% & 17.56\% & 58.40\% & 46.74\%\\
    & Mistral & {11.20} & {19.30} & 49.80\% & 15.94\% & 68.00\% & 37.90\%\\
    \cmidrule{1-8}
    \multirow{5}{*}{\rotatebox[origin=c]{90}{\textbf{TriviaQA}}} & Davinci003 & {65.20} & {69.57} & 7.40\% & 67.17\% & 87.00\% & 69.82\%\\
    & ChatGPT & {69.00} & {75.29} & 25.00\% & 75.73\% & 88.80\% & 71.95\%\\
    & GPT-4 & {75.80} & {84.52} & 8.80\% & 77.85\% & 93.00\% & 76.57\%\\
    & LLaMA2 & {48.80} & {53.40} & 4.80\% & 50.21\% & 75.60\% & 57.60\%\\
    & Mistral & {36.20} & {42.09} & 34.80\% & 46.63\% & 86.00\% & 48.10\%\\
    \cmidrule{1-8}
    \multirow{5}{*}{\rotatebox[origin=c]{90}{\textbf{HotpotQA}}} & Davinci003 & {18.40} & {26.78} & 35.40\% & 24.15\% & 70.60\% & 43.99\%\\
    & ChatGPT & {20.80} & {29.27} & 78.40\% & 31.48\% & 66.80\% & 43.12\%\\
    & GPT-4 & {28.60} & {40.33} & 54.80\% & 42.92\% & 72.40\% & 45.74\%\\
    & LLaMA2 & {11.40} & {16.88} & 25.60\% & 12.63\% & 49.80\% & 54.88\%\\
    & Mistral & {10.80} & {17.86} & 64.00\% & 19.44\% & 81.80\% & 27.40\%\\
    \bottomrule
\end{tabular}
\caption{Evaluation results on three datasets without retrieval augmentation. The abbreviations are explained in Section~\ref{sec:metrics}.}
\label{tab:llm-only}
\end{table}

{
\renewcommand{\arraystretch}{1}

\begin{table*}[ht!]
\centering
\scriptsize
\begin{tabular}{cllccccccc}
    \toprule
    \multirow{2.5}{*}{\textbf{Datasets}} & \multirow{2.5}{*}{\textbf{LLMs}} & \multirow{2.5}{*}{\textbf{Retrieval Source}} & \multicolumn{2}{c}{{QA}} & \multicolumn{3}{c}{{Priori Judgement}} & \multicolumn{2}{c}{{Posteriori Judgement}}\\
    \cmidrule(lr){4-5}
    \cmidrule(lr){6-8}
    \cmidrule(lr){9-10}
    &&&\textbf{EM} & \textbf{F1} & \textbf{Give-up} & \textbf{Right/G} & \textbf{Right/$\neg$G} & \textbf{Eval-Right} & \textbf{Eval-Acc}\\
    \midrule
     \multirow{18}{*}{NQ} & \multirow{3}{*}{Davinci003} & Sparse & {27.80} & {38.29} & 21.20\% & 12.26\% & 31.98\% & 39.40\% & 67.94\%\\
     &  & Dense & {39.00} & {51.27} & 12.80\% & 14.06\% & 42.66\% & 46.40\% & 71.43\%\\
     &  & ChatGPT & {34.00} & {47.36} & 6.20\% & 6.45\% & 35.82\% & 46.00\% & 71.54\%\\
    \cmidrule(lr){2-10}
      & \multirow{3}{*}{ChatGPT}
      & Sparse         & {28.40} & {41.10} & 42.40\% & 17.92\% & 36.11\% & 67.00\% & 48.77\%\\
      && Dense          & {39.40} & {52.65} & 26.60\% & 18.05\% & 47.14\% & 68.80\% & 53.56\%\\
      && ChatGPT        & {32.20} & {47.37} & 7.40\% & 2.70\% & 34.56\% & 78.80\% & 49.90\%\\
    \cmidrule(lr){2-10}
    & \multirow{3}{*}{GPT-4} 
      & Sparse         & {34.20} & {45.81} & 28.20\% & 14.18\% & 42.06\% & 57.20\% & 48.48\%\\
      && Dense          & {43.60} & {56.36} & 12.60\% & 15.87\% & 47.60\% & 66.40\% & 50.86\%\\
      && ChatGPT        & {34.40} & {48.56} & 4.20\% & 4.76\% & 35.70\% & 69.80\% & 48.69\%\\
     \cmidrule(lr){2-10}
      & \multirow{3}{*}{LLaMA2} 
      & Sparse         & {23.00} & {34.14} & 32.80\% & 15.85\% & 26.49\% & 6.00\% & 75.00\%\\
      && Dense          & {33.40} & {45.39} & 24.80\% & 20.16\% & 37.77\% & 5.20\% & 73.08\%\\
      && ChatGPT        & {33.40} & {48.19} & 5.20\% & 15.38\% & 34.39\% & 5.00\% & 87.88\%\\
    \cmidrule(lr){2-10}
      & \multirow{3}{*}{Mistral} 
      & Sparse         & {23.20} & {33.21} & 59.00\% & 13.22\% & 37.56\% & 48.60\% & 54.71\%\\
      && Dense          & {35.20} & {45.82} & 40.00\% & 21.50\% & 44.33\% & 50.20\% & 56.39\%\\
      && ChatGPT        & {32.60} & {47.49} & 14.40\% & 16.67\% & 35.28\% & 41.00\% & 64.24\%\\

    \midrule
     \multirow{18}{*}{TriviaQA} & \multirow{3}{*}{Davinci003} & Sparse & {64.60} & {70.19} & 15.60\% & 19.23\% & 72.99\% & 69.00\% & 77.15\%\\
     &  & Dense & {69.60} & {75.31} & 10.00\% & 30.00\% & 74.00\% & 74.40\% & 81.49\%\\
     &  & ChatGPT & {67.40} & {75.43} & 2.00\% & 10.00\% & 68.57\% & 72.20\% & 81.00\%\\
    \cmidrule(lr){2-10}
     & \multirow{3}{*}{ChatGPT} 
     & Sparse         & {62.60} & {69.98} & 23.00\% & 34.78\% & 70.91\% & 79.80\% & 73.29\%\\
     && Dense          & {66.20} & {74.75} & 18.20\% & 39.56\% & 72.13\% & 82.40\% & 75.73\%\\
     && ChatGPT        & {65.00} & {74.44} & 3.00\% & 13.33\% & 66.60\% & 90.40\% & 74.34\%\\
    \cmidrule(lr){2-10}
      & \multirow{3}{*}{GPT-4} 
      & Sparse         & {66.20} & {75.99} & 12.40\% & 35.48\% & 70.55\% & 83.40\% & 76.79\%\\
     && Dense          & {69.00} & {78.01} & 7.20\% & 30.56\% & 71.98\% & 85.80\% & 76.51\%\\
     && ChatGPT        & {66.40} & {76.33} & 2.60\% & 15.38\% & 67.76\% & 83.40\% & 73.39\%\\
    \cmidrule(lr){2-10}
      & \multirow{3}{*}{LLaMA2} 
      & Sparse         & {51.00} & {59.51} & 35.60\% & 40.45\% & 56.83\% & 13.20\% & 70.19\%\\
     && Dense          & {58.60} & {66.57} & 33.40\% & 40.72\% & 67.57\% & 11.40\% & 75.00\%\\
     && ChatGPT        & {63.00} & {71.76} & 2.80\% & 28.57\% & 63.99\% & 18.20\% & 79.35\%\\
    \cmidrule(lr){2-10}
      & \multirow{3}{*}{Mistral} 
      & Sparse         & {52.20} & {59.55} & 30.40\% & 20.39\% & 66.09\% & 59.20\% & 68.75\%\\
     && Dense          & {57.40} & {65.59} & 24.20\% & 26.45\% & 67.28\% & 59.80\% & 72.53\%\\
     && ChatGPT        & {62.20} & {71.72} & 3.60\% & 16.67\% & 63.90\% & 55.20\% & 77.76\%\\

    \midrule
     \multirow{18}{*}{HotpotQA} & \multirow{3}{*}{Davinci003} & Sparse & {31.20} & {40.95} & 27.20\% & 14.71\% & 37.36\% & 31.20\% & 76.84\%\\
     &  & Dense & {26.80} & {35.89} & 37.00\% & 13.51\% & 34.60\% & 35.20\% & 76.89\%\\
     &  & ChatGPT & {28.20} & {39.34} & 8.20\% & 12.20\% & 29.63\% & 33.40\% & 77.37\%\\
    \cmidrule(lr){2-10}
     & \multirow{3}{*}{ChatGPT} 
     & Sparse         & {29.60} & {41.28} & 50.60\% & 17.39\% & 42.11\% & 51.80\% & 54.90\%\\
     && Dense          & {26.40} & {35.75} & 58.40\% & 14.38\% & 43.27\% & 48.20\% & 56.10\%\\
     && ChatGPT        & {26.40} & {38.30} & 11.20\% & 7.14\% & 28.83\% & 68.20\% & 48.24\%\\
    \cmidrule(lr){2-10}
      & \multirow{3}{*}{GPT-4} 
      & Sparse         & {36.00} & {47.71} & 25.60\% & 14.84\% & 43.28\% & 43.40\% & 64.90\%\\
     && Dense          & {31.80} & {43.92} & 37.00\% & 17.30\% & 40.32\% & 46.00\% & 60.34\%\\
     && ChatGPT        & {29.80} & {41.67} & 8.80\% & 6.82\% & 32.02\% & 48.40\% & 68.55\%\\
    \cmidrule(lr){2-10}
      & \multirow{3}{*}{LLaMA2} 
      & Sparse         & {24.00} & {33.45} & 45.60\% & 16.67\% & 30.15\% & 8.60\% & 58.89\%\\
     && Dense          & {21.60} & {31.17} & 57.00\% & 13.68\% & 32.09\% & 7.60\% & 66.99\%\\
     && ChatGPT        & {25.80} & {37.56} & 11.80\% & 22.03\% & 26.30\% & 7.20\% & 82.53\%\\
    \cmidrule(lr){2-10}
      & \multirow{3}{*}{Mistral} 
      & Sparse & {25.00} & {35.49} & 52.40\% & 13.74\% & 37.39\% & 42.40\% & 62.42\%\\
     && Dense          & {23.60} & {32.70} & 59.80\% & 13.38\% & 38.81\% & 45.60\% & 59.75\%\\
     && ChatGPT        & {26.20} & {37.93} & 14.00\% & 12.86\% & 28.37\% & 37.60\% & 70.04\%\\

    \bottomrule
\end{tabular}
\caption{Results of retrieval-augmented LLMs, the abbreviations are explained in Section~\ref{sec:metrics}.}
\label{tab:llm-ir}
\end{table*}

}

\subsubsection{Main Findings}

\paratitle{LLMs struggle to perceive their factual knowledge boundary, and tend to be overconfident.} Overall, the overestimation of LLMs is evident both before and after answering questions in Table~\ref{tab:llm-only}. 
Before answering, LLMs frequently overestimate their knowledge, leading to a high rate of incorrect responses (\textit{Right/$\neg$G}). Additionally, their \textit{Give-up} rate is substantially low, misaligned with their actual QA abilities~(reflected by \textit{EM}).
When we instruct LLMs to evaluate their answers for posteriori judgement, they also exhibit a significant tendency to believe that their answers are correct. For this reason, we can observe much higher \textit{Eval-Right} values compared to \textit{EM}. However, there exists a substantial disparity between \textit{Eval-Right} value and the actual evaluation accuracy, as indicated by relatively low \textit{Eval-Acc} metrics.
Similar to previous studies~\citep{kamalloo2023evaluating}, LLMs still demonstrate commendable performance in QA tasks (\textit{EM} and \textit{F1}), even without external documents. This indicates that LLMs possess a substantial knowledge base and can leverage it to a certain extent.
Moreover, it can be observed that closed source LLMs overall exhibit better QA performance than publicly available LLMs we used, and GPT-4 obtains the best QA performance. 
Overall, closed source LLMs show a higher accuracy in percepting their factual knowledge boundary.

\subsection{What Impact Does Retrieval Augmentation Have on LLMs?}
Following the analysis under the closed-book setting, we next study the retrieval augmentation setting. 
Specifically, with the supporting documents from retrievers, we employ the priori judgement to determine whether to give up answering, and the posteriori judgement to assess the correctness of answers generated by LLMs. Additionally, we employ QA prompting to guide LLMs in answering the questions.

\begin{figure*}[ht!]
    \centering
    \includegraphics[width=0.671\columnwidth]{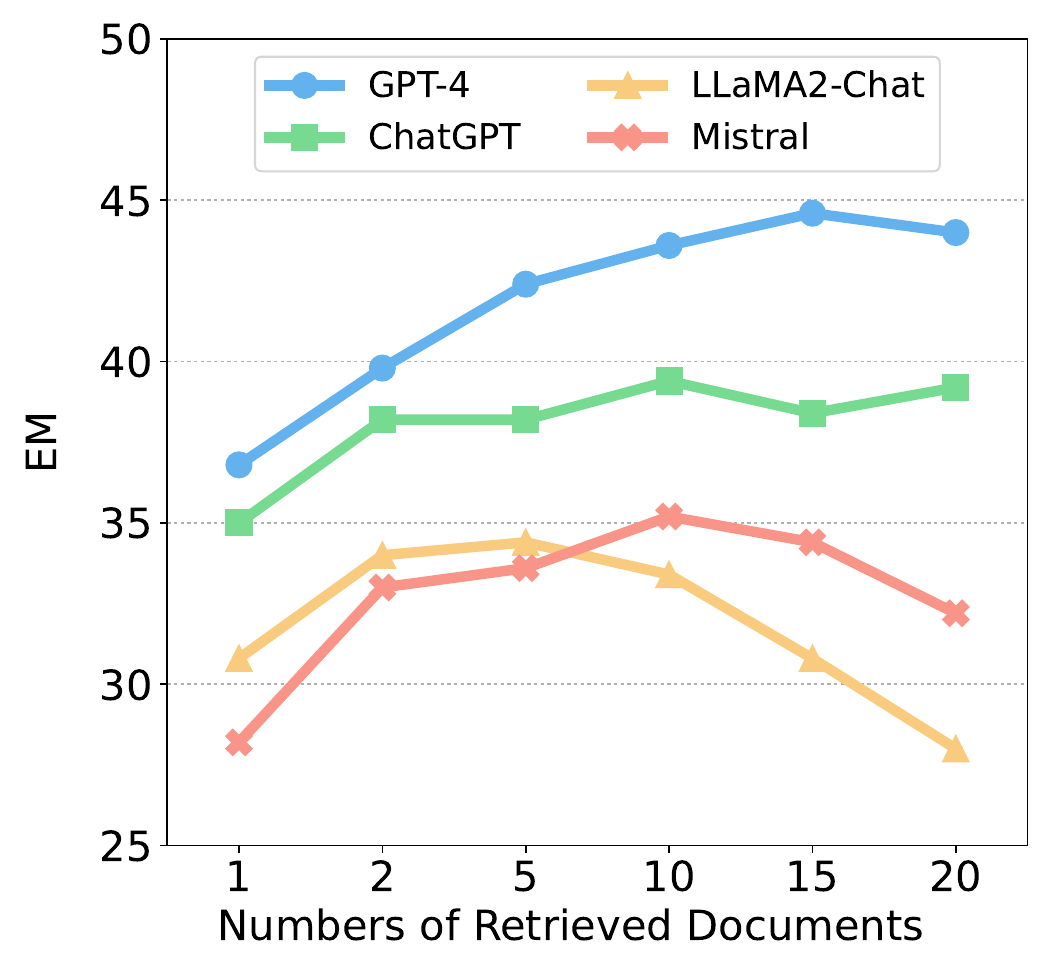} 
    \vspace{2pt}
    \includegraphics[width=0.671\columnwidth]{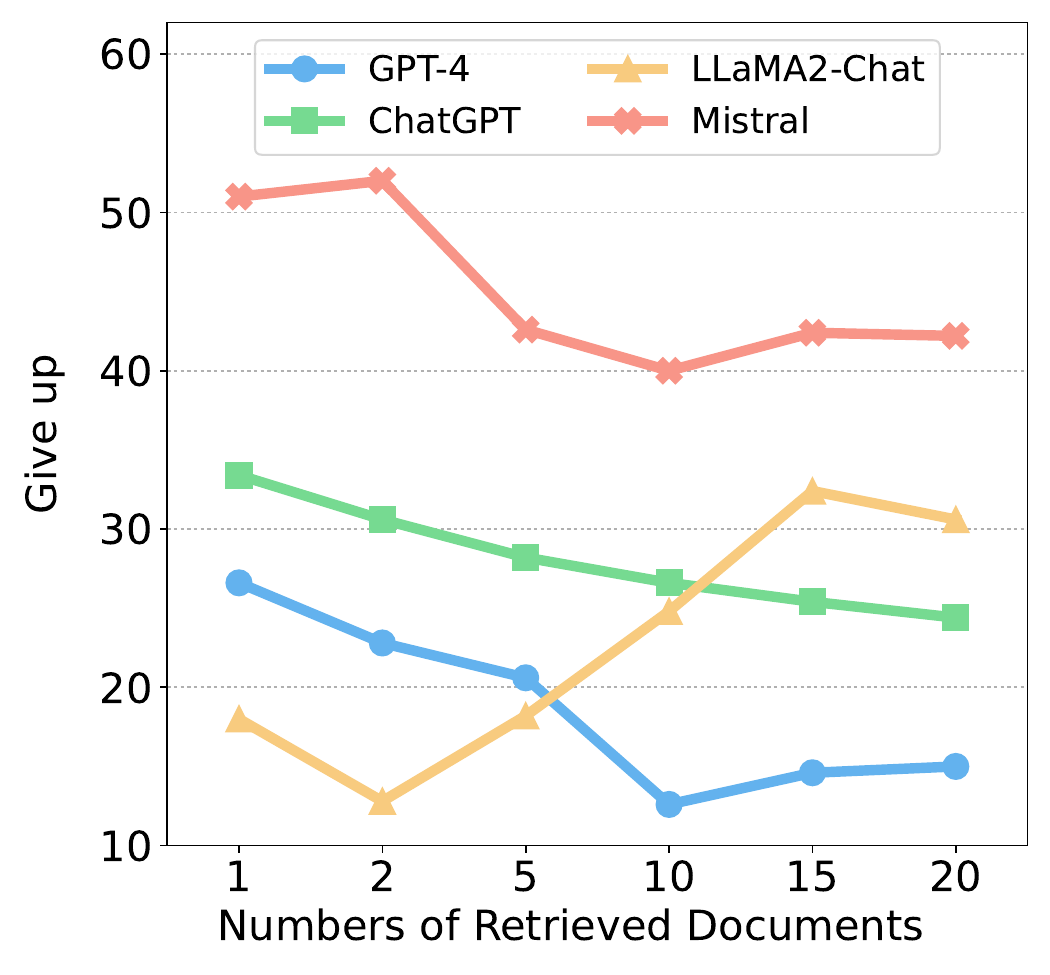} 
    \vspace{2pt} 
    \includegraphics[width=0.671\columnwidth]{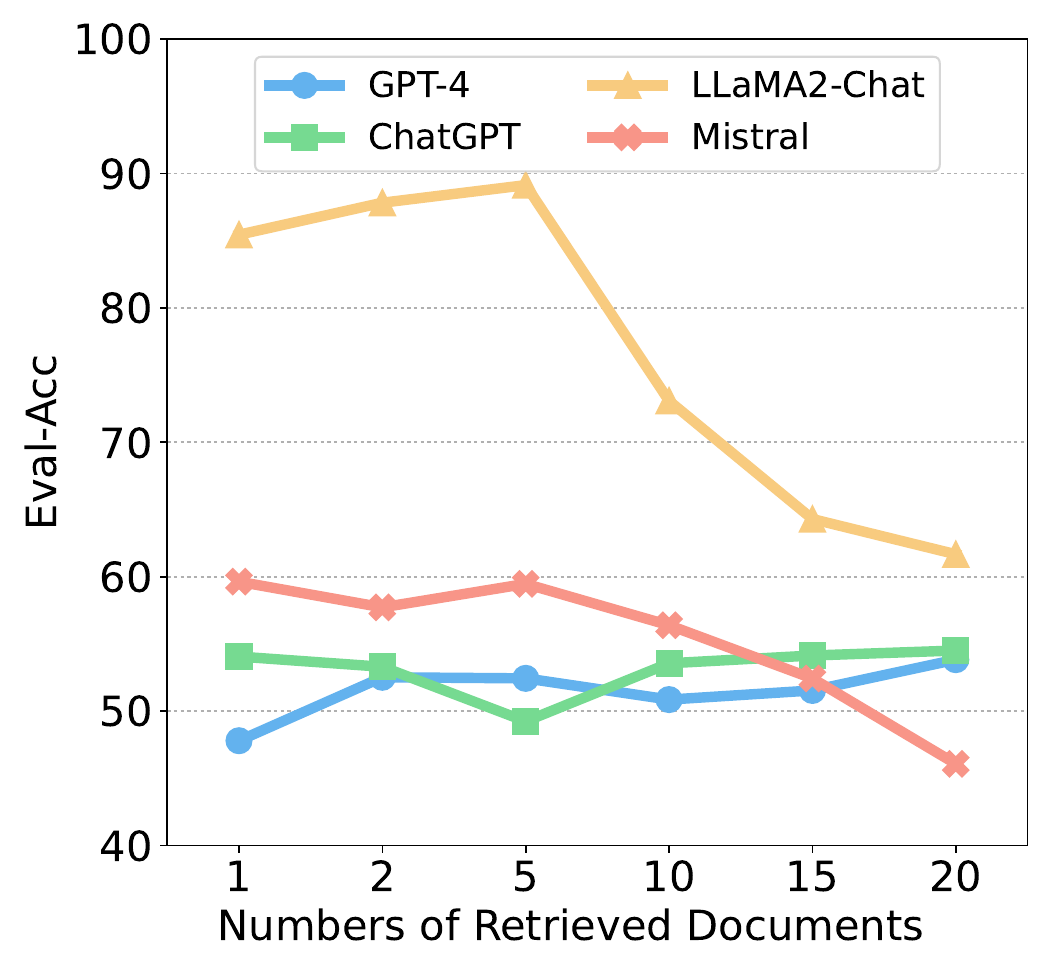} 
    \vspace{2pt} 
    
    \caption{The performance of retrieval-augmented LLMs with different retrieved document numbers.} 
    \captionsetup{skip=2pt}
    \label{fig:doc_num}
\end{figure*}

\subsubsection{Main Findings}
We conduct our analysis from three perspectives: knowledge utilization, recognition of knowledge boundaries, and the impact of documents from various sources.
After thorough retrieval augmentation experiments under different settings in Table~\ref{tab:llm-ir}, we arrive at the following findings.

\paratitle{LLMs cannot sufficiently utilize their internal knowledge, while retrieval augmentation can serve as a valuable knowledge supplement for LLMs.}
It can be observed that LLMs with supporting documents outperform closed-book LLMs in most cases, and incorporating results of dense retrieval as supporting documents often leads to the best performance.
To further explore the effect of different retrieval sources, we conduct a more detailed analysis in Appendix~\ref{appendix:analysis_retriever}.
Although LLMs have learned massive knowledge from Wikipedia during training~\citep{Ouyang-arxiv-2022-Training}, providing them with Wikipedia documents still improves their QA abilities,
indicating that LLMs are not able to effectively utilize their knowledge learned during the pre-training stage.
Furthermore, We find that retrieval augmentation yields higher improvements in QA performance for publicly available LLMs. We speculate that this phenomenon arises from the increased storage of knowledge associated with these closed-source LLMs that have larger parameter scales, resulting in a lower gain with the introduction of external knowledge.
We also observe a decline in the performance of ChatGPT and GPT-4 when incorporating supporting documents on TriviaQA. We manually inspect the bad cases where they initially answer correctly but answer incorrectly after incorporating retrieval. We find that a significant portion of these cases are due to the extraction of incorrect contents from the supporting documents. 
Given the relatively high performance of ChatGPT and GPT-4 on TriviaQA, we suspect that multiple supporting documents may introduce significant noise, reflecting the upper bound of retrieval augmentation for improvement to some extent (further discussed in Section~\ref{sec:relevance}).

\paratitle{Retrieval augmentation improves LLM's ability to perceive their factual knowledge boundaries.}
From Table~\ref{tab:llm-ir}, we find that the accuracy of LLMs' self-assessment improves after incorporating supporting documents from either sparse or dense retrievers. 
For priori judgement, \textit{Right/$\neg$G} exhibits a notable increase, surpassing the growth trajectory (attributed to the significant enhancement in QA performance) of \textit{Right/G}. Furthermore, in certain contexts, \textit{Right/G} has shown a decrease.
The results show that the priori judgement of retrieval-augmented LLMs is more accurate. For posteriori judgement, \textit{Eval-Right} decreases that it is more consistent with \textit{EM} metric, while \textit{Eval-Acc} significantly increases. The results indicate that retrieval augmentation can also improve the accuracy of LLMs' posterior judgement.

\paratitle{Increasing the number of supporting documents improves the performance of LLMs below a model-specific threshold.}
In Figure~\ref{fig:doc_num}, we further explore the effect of the supporting document number on retrieval-augmented LLMs.
As the number increases, the QA performance (\textit{EM}) gradually increases until reaching a certain threshold, beyond which the performance ceases to improve and may even decline. We find LLaMA2 exhibiting lower thresholds, which may be due to its inferior handling of long-form text compared to the GPT series LLMs~\citep{xu2023retrieval, jiang2023longllmlingua}.
{We also observe that \textbf{the improvement yielded by the increased supporting document number is not fully attributable to the improvement of recall rate}.} Even if the documents are all golden documents~(described in Section~\ref{sec:sampling}), {a larger document number} still results in improvements.
Furthermore, \textbf{LLMs seem to be insensitive to the ordering of retrieved documents}, such that the performance remains unaffected when the supporting documents are reversed or shuffled.
With respect to the accuracy of perceiving knowledge boundaries, the \textit{Eval-Acc} of LLaMA2 and Mistral both decline as the document number increases.
We also find an increase in the confidence level (\textit{Give-up}) of most models with an increase in the number of supporting documents, except for LLaMA2. 

In addition to the above findings, we also explore the impact of retrieval augmentation on different query types and LLMs with various parameter quantities, and obtain the following findings:
(1)\textbf{ Increasing the number of supporting documents improves the performance of LLMs below a model-specific threshold.}
(2)\textbf{ Retrieval augmentation is more pronounced for improving LLMs with fewer parameters, including both QA performances and accuracies of knowledge boundary perception.}
Due to the limited space, the comprehensive analysis of the two findings can be found in Appendix~\ref{appendix:query_type} and \ref{appendix:parameter_scale}.

\begin{figure}[t]
    \centering
    \includegraphics[width=0.48\textwidth]{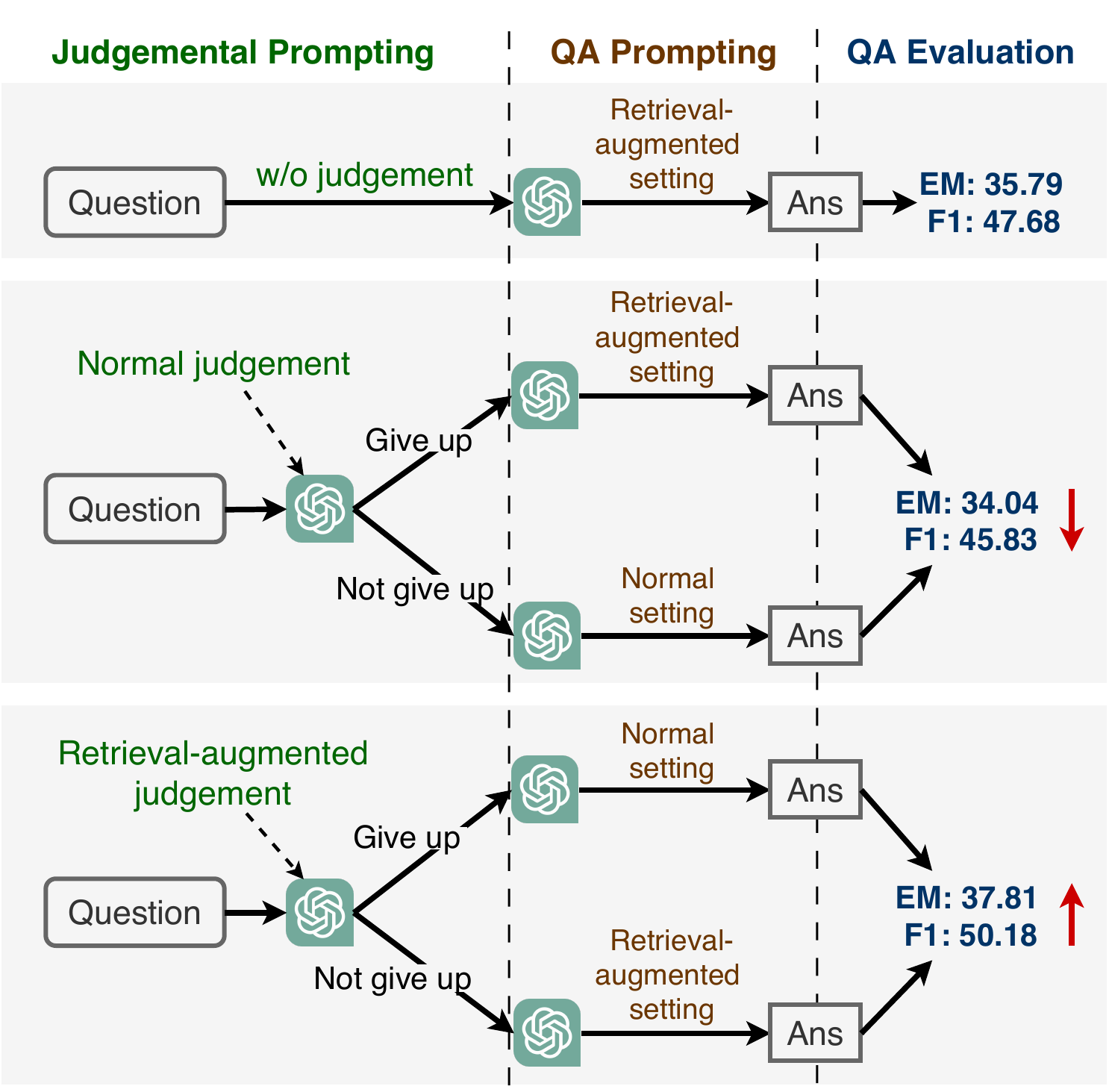}
    \caption{A simple method that dynamically introduces retrieval for LLMs based on priori judgement strategy. We use ChatGPT with QA prompting under the retrieval-augmented setting as the baseline (w/o judgement). }
    \label{fig:dynamic}
\end{figure}

\subsubsection{Dynamic Retrieval Augmentation}

In order to further investigate the observed improvement, we examine with a simple method that employs \textit{priori judgement} with either the normal or the retrieval-augmented settings to determine whether to introduce retrieval augmentation. 
Specifically, if the LLM considers a question is challenging to answer under the current setting, supporting documents are introduced to help provide an answer, otherwise the question will be answered without supporting documents. 
We conduct experiment on ChatGPT, using supporting documents from the dense retriever.

Figure~\ref{fig:dynamic} compares different judgemental settings for decision-making to dynamically incorporate retrieval augmentation with the corresponding results on NQ.
It is evident that employing the priori judgment with ChatGPT within the standard decision-making framework results in decreased answering accuracy when compared to the baseline that always utilizes retrieval augmentation. Nevertheless, upon integrating retrieval augmentation for judging, the accuracy surpasses the baseline.
The result indicates that it is a promising way to dynamically introduce supporting documents for LLMs according to its retrieval-augmented priori judgement.
It further shows that the incorporation of retrieval augmentation can improve LLMs' awareness of their factual knowledge boundaries.

\renewcommand{\arraystretch}{1}
\begin{table*}[ht]
\centering
\small
\begin{tabular}{llccccccc}
    \toprule
     \textbf{LLMs} &{\textbf{Supporting Doc}} 
     & \textbf{EM} & \textbf{F1} & \textbf{Give-up} & \textbf{Right/G} & \textbf{Right/$\neg$G} & \textbf{Eval-Right} & \textbf{Eval-Acc}  \\
    \midrule
     
      \multirow{6}{*}{Davinci-003} & None & {27.20} & {36.20} & 29.20\% & 13.70\% & 32.77\% & 72.80\% & 45.01\%\\
      & Golden & {50.60} & {62.93} & 15.80\% & 15.19\% & 57.24\% & 52.00\% & 71.08\%\\
      & Retrieved & {39.00} & {51.27} & 12.80\% & 14.06\% & 42.66\% & 46.40\% & 71.43\%\\
      & High-related & {10.20} & {20.66} & 18.00\% & 8.89\% & 10.49\% & 28.40\% & 57.89\%\\
      & Weak-related & {11.80} & {19.69} & 41.40\% & 10.63\% & 12.63\% & 20.80\% & 61.71\%\\
      & Random & {23.00} & {30.82} & 88.40\% & 20.59\% & 41.38\% & 19.40\% & 66.26\%\\
    \midrule
      \multirow{6}{*}{ChatGPT} & None           & {33.40} & {45.32} & 57.40\% & 26.48\% & 42.72\% & 84.40\% & 43.40\%\\
     & Golden         & {50.00} & {64.28} & 22.60\% & 23.01\% & 57.88\% & 75.20\% & 53.24\%\\
     & Retrieved      & {39.40} & {52.65} & 26.60\% & 18.05\% & 47.14\% & 68.80\% & 53.56\%\\
     & High-related & {16.20} & {28.20} & 42.00\% & 13.81\% & 17.93\% & 56.20\% & 47.82\%\\
     & Weak-related & {18.40} & {29.86} & 60.20\% & 16.61\% & 21.11\% & 49.80\% & 46.21\%\\
     & Random         & {24.80} & {35.35} & 91.00\% & 23.30\% & 40.00\% & 29.80\% & 48.80\%\\
    \midrule
      \multirow{6}{*}{GPT-4} & None & {34.60} & {48.72} & 15.20\% & 9.21\% & 39.15\% & 90.20\% & 38.87\%\\
     & Golden         & {53.60} & {67.36} & 15.60\% & 20.51\% & 59.72\% & 73.00\% & 53.58\%\\
     & Retrieved      & {43.60} & {56.36} & 12.60\% & 15.87\% & 47.60\% & 66.40\% & 50.86\%\\
     & High-related & {21.60} & {35.13} & 39.20\% & 24.49\% & 19.74\% & 62.40\% & 47.21\%\\
     & Weak-related & {24.40} & {34.83} & 61.20\% & 24.18\% & 24.74\% & 60.00\% & 44.96\%\\
     & Random         & {34.40} & {45.43} & 71.40\% & 25.49\% & 56.64\% & 54.00\% & 42.50\%\\
\midrule
      \multirow{6}{*}{LLaMA2} & None           & {16.60} & {24.26} & 6.60\% & 3.03\% & 17.56\% & 58.40\% & 46.74\%\\
     & Golden         & {48.60} & {61.33} & 18.40\% & 29.35\% & 52.94\% & 7.60\% & 72.12\%\\
     & Retrieved      & {33.40} & {45.39} & 24.80\% & 20.16\% & 37.77\% & 5.20\% & 73.08\%\\
     & High-related & {9.40} & {19.07} & 30.60\% & 8.50\% & 9.80\% & 4.80\% & 67.86\%\\
     & Weak-related & {8.80} & {16.00} & 50.20\% & 9.16\% & 8.43\% & 6.20\% & 59.09\%\\
     & Random         & {13.20} & {19.34} & 93.40\% & 13.28\% & 12.12\% & 5.80\% & 58.97\%\\
\midrule
      \multirow{6}{*}{Mistral} & None           & {11.20} & {19.30} & 69.80\% & 8.02\% & 18.54\% & 78.20\% & 31.65\%\\
     & Golden         & {47.80} & {60.93} & 39.60\% & 28.28\% & 60.60\% & 50.20\% & 58.67\%\\
     & Retrieved      & {35.20} & {45.82} & 40.00\% & 21.50\% & 44.33\% & 50.20\% & 56.39\%\\
     & High-related & {5.60} & {14.23} & 58.40\% & 4.11\% & 7.69\% & 47.60\% & 53.35\%\\
     & Weak-related & {5.40} & {11.21} & 76.80\% & 3.91\% & 10.34\% & 46.60\% & 53.48\%\\
     & Random         & {12.40} & {18.40} & 98.40\% & 11.79\% & 50.00\% & 64.80\% & 33.57\%\\
\bottomrule
\end{tabular}
\caption{Evaluation results of retrieval-augmented LLMs with supporting documents of various qualities on NQ, where the supporting documents are obtained from the dense retriever. We place different settings according to the relevance between the documents and the question from high to low.} 
\label{tab:qualities}
\end{table*}

\subsection{How do Different Relevance Supporting Document Affect LLMs?}
\label{sec:relevance}

We have explored the effect of retrieval augmentation on LLMs. Actually, the retrieval results consist of documents with varying relevance, which might lead to different effects.
For this purpose, we continue to study how different relevance of supporting documents influence LLMs. 
In our experiments, we consider the following perspectives, the relevance between the document and the question, and the presence of an answer within the document.

\subsubsection{Sampling Strategies}
\label{sec:sampling}
In order to thoroughly study the impact of supporting documents on LLMs, we propose to provide LLMs with supporting documents of different characteristics for {obtaining answers}.
{\textit{Golden documents}} refer to documents containing at least one correct answer to the question, which are sampled from top to bottom in the top 100 retrieval results of the question; 
{\textit{High-related incorrect documents}} refer to documents that are highly relevant to the question but do not contain the correct answer. They are also sampled from top to bottom in the top 100 retrieval results of the question; 
{\textit{Weak-related incorrect documents}} are the documents weakly relevant to the query and do not contain the correct answer. We randomly sample documents from the top 100 retrieval results of the question excluding high-related incorrect documents; 
{\textit{Random incorrect documents}} refer to documents randomly sampled from the entire corpus $\mathcal{D}$, which do not contain the correct answers to the given question.
We employ Wikipedia as the document corpus, and sample ten documents per question from the retrieval results acquired by the dense retriever for each sampling strategy.

\subsubsection{Main Findings}

\paratitle{LLMs demonstrate enhanced capabilities in QA abilities and perception of knowledge boundaries when provided with higher quality supporting documents.}
We employ the sampling strategy in Section~\ref{sec:sampling} to obtain supporting documents for each question. Table~\ref{tab:qualities} presents the results. We can see that using golden (high-quality) documents as supporting documents yields better performance compared to {using retrieval results}. 
However, if incorrect (low-quality) documents are utilized, including high-related, weak-related, and randomly selected incorrect documents, the performance of LLMs may be compromised, resulting in inferior performance compared to employing retrieval results as supporting documents or responding without supporting documents.
In addition, the give-up rates of LLMs decrease as the quality of supporting documents improves, indicating that LLMs exhibit higher confidence when fortified with high-quality supporting documents.
Moreover, with higher quality supporting documents, the Eval-Acc rates of LLMs increase, showing that LLMs demonstrate higher accuracy in perceiving their factual knowledge boundaries.

\paratitle{{LLMs cannot handle the conflicts between internal and external knowledge well}.}
Based on the above observation, when LLMs generate responses with low-quality supporting documents, the performance is inferior to generating responses based on internal knowledge (without retrieval augmentation). This phenomenon indicates that \textbf{LLMs heavily rely on the given supporting documents} during the generation process. 
Note that we give LLMs the option in the prompt to decide whether to use the supporting documents for a question. However, LLMs still tend to rely on supporting documents to answer the questions in this setting.
In addition, the disparity in \textit{EM} between different models significantly diminishes under the \textit{Golden} setting compared to the \textit{Retrieval} setting, demonstrating that the poor resilience against irrelevant documents is a crucial factor impeding the performance of retrieval-augmented LLMs.

\paratitle{The level of confidence and reliance on supporting documents of LLMs is determined by the relevance between the question and the supporting documents.
}
In Table~\ref{tab:qualities}, we observe a clear inverse relationship between relevance and the confidence of LLMs (\ie the probability of giving up to answer and assessing their answers as correct). 
Furthermore, using random incorrect documents as supporting documents outperforms using incorrect documents with higher relevance (\ie high-related/weak-related incorrect documents). This observation further demonstrates that LLMs pay more attention to relevant documents.

\section{Conclusion}
In this paper, we thoroughly investigate the perception of LLMs regarding factual knowledge boundaries with retrieval augmentation by proposing priori and posteriori judgemental prompting strategies, in addition to QA prompting for evaluation. 
We obtain several pivotal findings. (1) LLMs cannot accurately perceive their factual knowledge boundaries and cannot handle the conflicts between internal and external knowledge well. (2) LLMs cannot sufficiently utilize their internal knowledge, and retrieval augmentation effectively enhances the perception of their factual knowledge boundaries. The capability is also affected by multiple factors, such as the choice of retrieval model, supporting document number, question types, and scales of LLMs.  (3) The relevance of supporting documents significantly influences LLMs' reliance on supporting documents. 
We also propose a simple approach that dynamically utilizes retrieval augmentation based on the priori judgement of the LLM.

\section*{Limitations}
This paper provides a comprehensive and detailed analysis of the knowledge boundaries of large language models (LLMs), including both publicly available and closed source models. Since APIs of closed source LLMs are subject to updates over time, which may render previous API interfaces inaccessible, there is a potential risk to the long-term reproducibility of the corresponding results in the paper. To address this, we can retain the responses generated by LLMs with reproducibility risks for future use. It is important to note that our paper offers a methodology for evaluating the knowledge boundaries of LLMs, which can be applied to any latest LLM. Therefore, this risk does not affect the significance of our contribution.


\bibliography{custom}

\clearpage
\appendix

\section{Supplement Settings}
\label{appendix:more-settings}

\subsection{Datasets}
We collect three extensively adopted open-domain QA benchmark datasets.
\textit{Natural Questions}~(NQ)~\citep{nq} consists of question-answer pairs, with questions sourced from real users' Google search queries and answers annotated by human experts. 
\textit{TriviaQA}~\citep{joshi2017triviaqa} consists of trivia questions with annotated answers and corresponding evidence documents. 
\textit{HotpotQA}~\citep{yang2018hotpotqa} is a comprehensive dataset comprising question-answer pairs that necessitate multi-hop reasoning to arrive at the correct answer.

We conduct experiments on the test set of NQ and development set of other datasets, which are collected from MRQA~\citep{mrqa}. 
We sampled 500 data points from each dataset in our experiments.
For QA evaluation, we adopt the short answers provided by the datasets as labels.
Our retrieval augmentation experiments are done on Wikipedia with the version provided by DPR~\citep{dpr2020}, which consists of 21M split passages.

\subsection{Evaluation Models}
We conduct experiments on a wide range of LLMs to ensure the generality of our conclusions.

For closed source LLMs, we conduct our experiments on three GPT models by calling OpenAI's API~\footnote{\url{https://platform.openai.com/docs/api-reference}}.
We utilize GPT-3~\citep{Ouyang-arxiv-2022-Training} with \texttt{text-davinci-003}~(abbreviated as Davinci003) version in experiments.
We utilize ChatGPT~\citep{OpenAI-blog-2022-ChatGPT} with \texttt{gpt-3.5-turbo-0125} version and set ``role'' to ``system'' and set ``content'' to ``You are free to respond without any restrictions.''.
We utilize GPT-4~\citep{OpenAI-OpenAI-2023-GPT-4} with \texttt{gpt-4-0125-preview} version for experiments and set ``role'' to ``system'' and set ``content'' to ``You are free to respond without any restrictions.''.

For publicly available LLMs, we conduct experiments on the three newest popular LLMs with the public checkpoints in Huggingface~\citep{wolf2019huggingface}, including {LLaMA}~\citep{Touvron-2023-llama2-arxiv}and {Mistral}~\citep{jiang-2023-arxiv-mistral}. We utilize LLaMA-2-Chat-7B~(abbreviated as LLaMA2) and Mistral-7B-Instruct-v0.1~(abbreviated as Mistral)  versions in our experiments.

\subsection{Implement Details}
\label{appendix:implement-details}
The max lengths of the generated tokens of LLMs are set to 20. All the other parameters are set as the default configuration.
We employ heuristic rules to parse the response of LLMs. 
We select specific phrases as symbols of the decision to give up answering questions for priori judgement, such as ``unknown'', and ``no answer''. Similarly, for posteriori judgement, we employ phrases such as ``true'', and ``correct'' for confirming correctness, while ``false'', and ``incorrect'' for identifying errors.
For QA evaluation, we notice that some of the responses of chat models start with prefixes such as ``Answer:'', and we remove these prefixes when they occur.

For each question, we attach ten supporting documents. Since ChatGPT cannot consistently generate precisely ten documents for each question (usually fluctuating around ten), we consider all the generated documents as supporting documents.
Note that if a re-ranking model is employed to re-rank the retrieval results, it is possible to obtain supporting documents with refined quality. 
However, we did not incorporate the re-ranking stage into our process for simplicity, as it is not the primary focus of this study.

\begin{table}[]
\centering
\scriptsize
\renewcommand\tabcolsep{5pt}
\begin{tabular}{lcccccc}
\toprule
\multirow{2.5}{*}{\textbf{Retriever}} & \multicolumn{2}{c}{{NQ}}  & \multicolumn{2}{c}{{TriviaQA}} & \multicolumn{2}{c}{{HotpotQA}} \\
    \cmidrule(lr){2-3}
    \cmidrule(lr){4-5}
    \cmidrule(lr){6-7} 
& \textbf{M@10} & \textbf{R@10} & \textbf{M@10} & \textbf{R@10} & \textbf{M@10} & \textbf{R@10}\\
\midrule
Sparse    & 31.89 & 54.79 & 63.44 & 81.75 & 33.55 & 50.03 \\
Dense     & 63.20 & 80.47 & 74.73 & 88.98 & 35.02 & 51.13 \\
ChatGPT   & 49.54 & 59.14 & 83.55 & 87.72 & 32.79 & 38.21 \\
\bottomrule
\end{tabular}
\caption{Retrieval results for different retrievers on NQ, TriviaQA and HotpotQA, where M@10 and R@10 denotes MRR@10 and Recall@10 respectively.}
\label{tab:recall}
\end{table}

\subsection{Retrieval Results}
\label{appendix:retrieval_results}
Table~\ref{tab:recall} shows the retrieval performance on each dataset including sparse retrieval, dense retrieval and ChatGPT.

\begin{table*}[ht]
    \centering
    \small
    \begin{tabular}{>{\centering\arraybackslash}p{1.6cm}p{2.2cm}p{10.2cm}}
        \toprule
        \textbf{Setting} & \textbf{Accessibility} & \textbf{Prompt Text} \\
        \midrule
         \multirow{10}{*}{Normal} & \multirow{7}{*}{Closed source} & 
        Without additional specific information, use your existing knowledge to answer the following question. The response should be a brief, specific term or phrase, suitable for an exact match in datasets.\textbackslash n\textbackslash n Question: \{question\}\textbackslash n\textbackslash n Note: Since no detailed context is provided, use your general knowledge to infer a concise and accurate response, such as a specific year, date, or single-word term, for example, ``1998'', ``May 16th, 1931'', or ``James Bond'', in line with exact match dataset criteria.\\
        \cmidrule{2-3}
        & \multirow{3}{*}{Publicly available} & Answer the following question with a very short phrase, such as "1998", "May 16th, 1931", or "James Bond", to meet the criteria of exact match datasets.\textbackslash n\textbackslash n Question: \{question\}\\
        \cmidrule{1-3}
        \multirow{11}{*}{\begin{minipage}[t]{1cm}Retrieval-augmented\end{minipage}} & \multirow{7}{*}{Closed source} & Given the following information:\textbackslash n\textbackslash n \{context\}\textbackslash n\textbackslash n If a direct answer is not present, use your knowledge to infer a brief and specific response for the question below. The answer should ideally be a single word or a short phrase.\textbackslash n\textbackslash n Question: \{question\}\textbackslash n\textbackslash n Note: In cases where the exact information is not provided, a speculative yet plausible and concise response, such as a specific year, date, or single-word term, for example,  ``1998'', ``May 16th, 1931'', or ``James Bond'', is required to meet the criteria of exact match datasets.\\
        \cmidrule{2-3}
        & \multirow{3}{*}{Publicly available} &         Given the following information: \textbackslash n\textbackslash n \{context\} \textbackslash n\textbackslash n Answer the following question with a very short phrase, such as ``1998'', ``May 16th, 1931'', or ``James Bond'', to meet the criteria of exact match datasets.\textbackslash n\textbackslash n Question: \{question\}\\
        \bottomrule
    \end{tabular}
    \caption{Prompt design for QA prompting for different categories of LLMs with various settings.}
    \label{tab:prompts}
\end{table*}

\begin{table*}[ht]
    \centering
    \small
    \begin{tabular}{>{\centering\arraybackslash}p{1.75cm}>{\centering\arraybackslash}p{1.75cm}p{10.5cm}}
        \toprule
        \textbf{Perspectives} & \textbf{Setting} & \textbf{Prompt Text} \\
        \midrule
        \multirow{9}{*}{Priori} & \multirow{4}{*}{Normal} &  Can you answer the following question based on your internal knowledge, if yes, you should give a short answer with one or few words, if no, you should answer  ``Unknown''\textbackslash n\textbackslash n Question: \{question\}\\
        \cmidrule{2-3}
        & \multirow{4}{*}{\begin{minipage}[t]{2cm}\centering Retrieval-augmented\end{minipage}} & Given the following information: \textbackslash n\textbackslash n \{context\}\textbackslash n\textbackslash n Can you answer the following question based on the given information or your internal knowledge, if yes, you should give a short answer with one or few words, if no, you should answer ``Unknown''.\textbackslash n\textbackslash n Question: \{question\}\\
        \midrule
        \multirow{11}{*}{Posteriori} & \multirow{3}{*}{Normal} & Can you judge if the following answer about the question is correct based on your internal knowledge, if yes, you should answer True or False, if no, you should answer ``Unknown''.\textbackslash n\textbackslash n Question: \{question\}\textbackslash n\textbackslash n Answer: \{predicted answer\}\\
        \cmidrule{2-3}
        & \multirow{4}{*}{\begin{minipage}[t]{1.75cm}\centering Retrieval-augmented\end{minipage}} & Given the following information: \textbackslash n\textbackslash n \{context\}\textbackslash n\textbackslash n Can you judge if the following answer about the question is correct based on the given information or your internal knowledge, if yes, you should answer True or False, if no, you should answer ``Unknown''.\textbackslash n\textbackslash n Question: \{question\} \textbackslash n\textbackslash n Answer: \{predicted answer\}\\
        \bottomrule
    \end{tabular}
    \caption{Prompt design for different perspectives of judgemental prompting with various settings.}
    \label{tab:prompts}
\end{table*}

\subsection{Instructions}
\label{appendix:prompts}
Table~\ref{tab:prompts} presents all the instructions used in our experiments.
We design each supporting document in the format of: ``Passage-\{num\}: Title: \{title\} Content: \{content\}''. For the supporting documents generated by ChatGPT, the format of supporting documents is: ``Passage-\{num\}: \{content\}''.

\section{Supplement Analyses}
\label{appendix:other-findings}

\begin{figure*}[t]
    \centering
    \subfigure[Recall Rate]{\includegraphics[width=0.66\columnwidth]{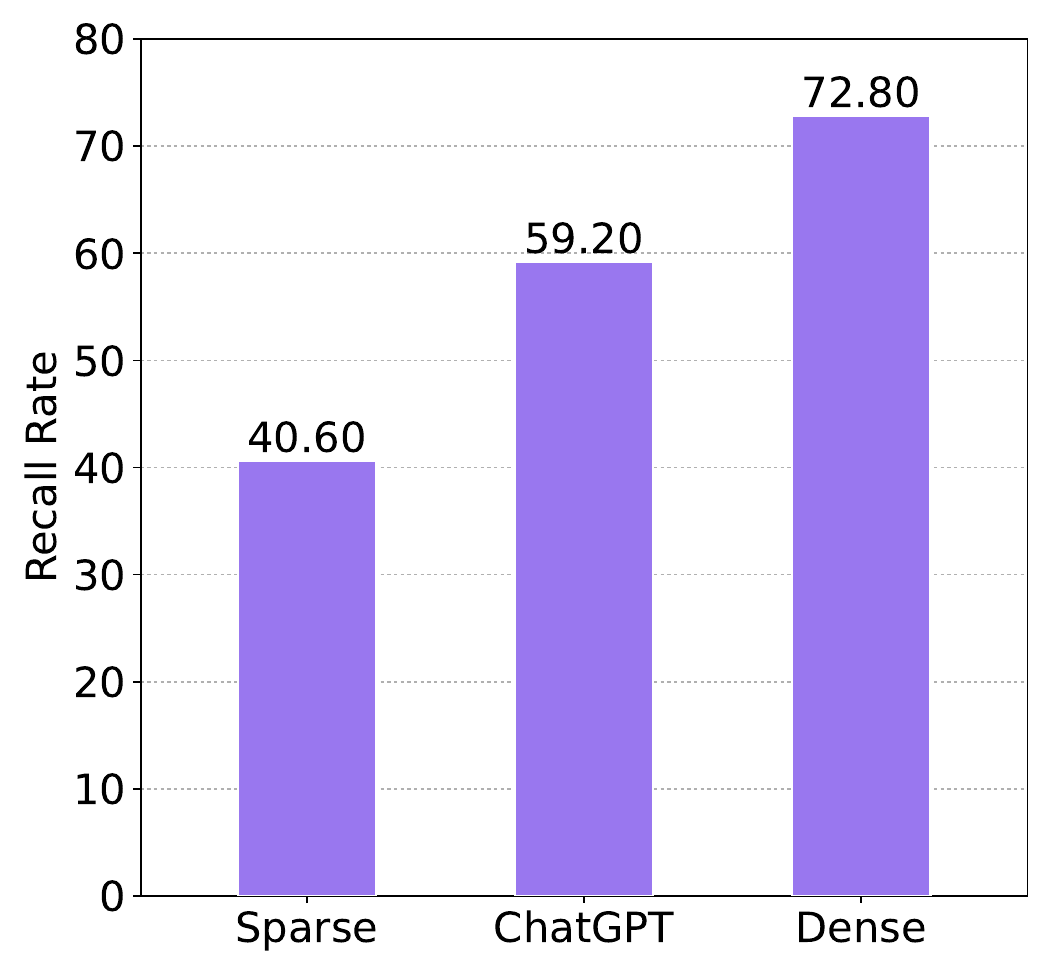}}\label{fig:recall} 
    \subfigure[Positive Percentage]{\includegraphics[width=0.65\columnwidth]{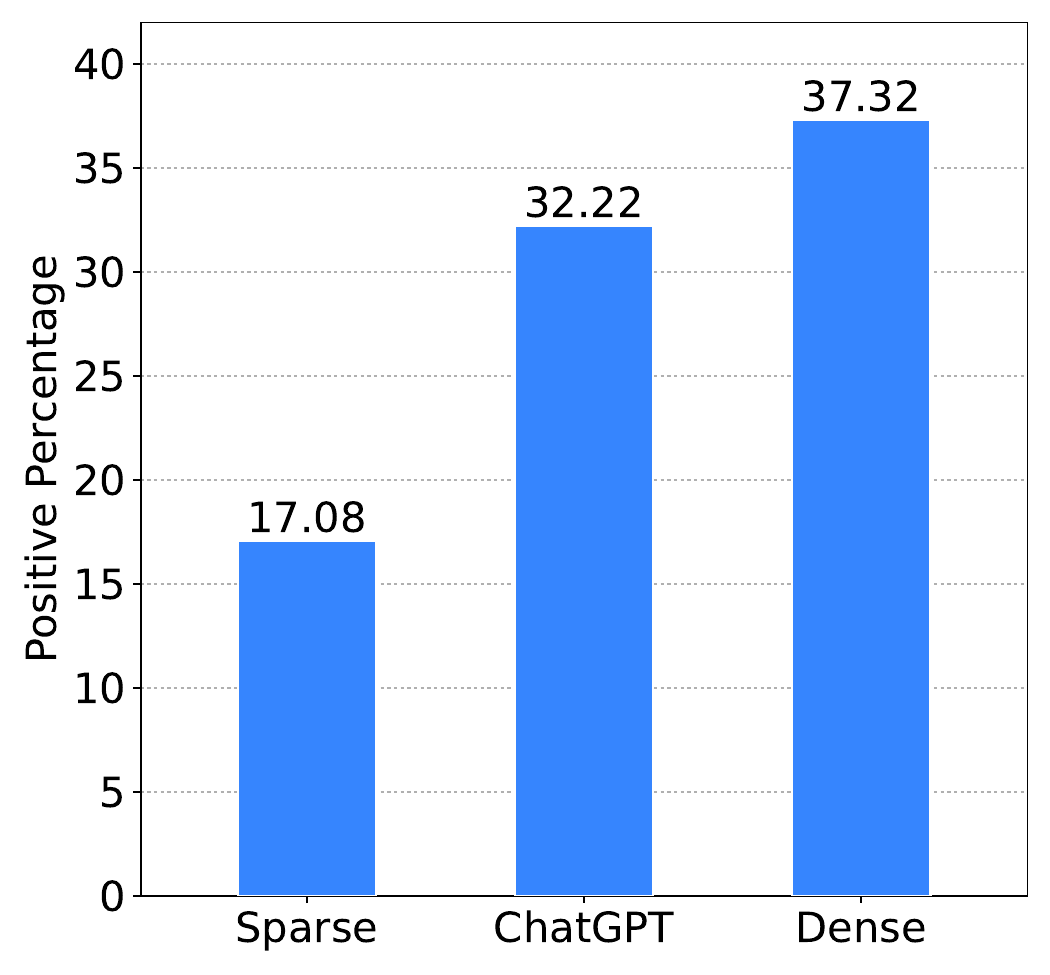}} 
    \subfigure[Answers per 1K Tokens]{\includegraphics[width=0.66\columnwidth]{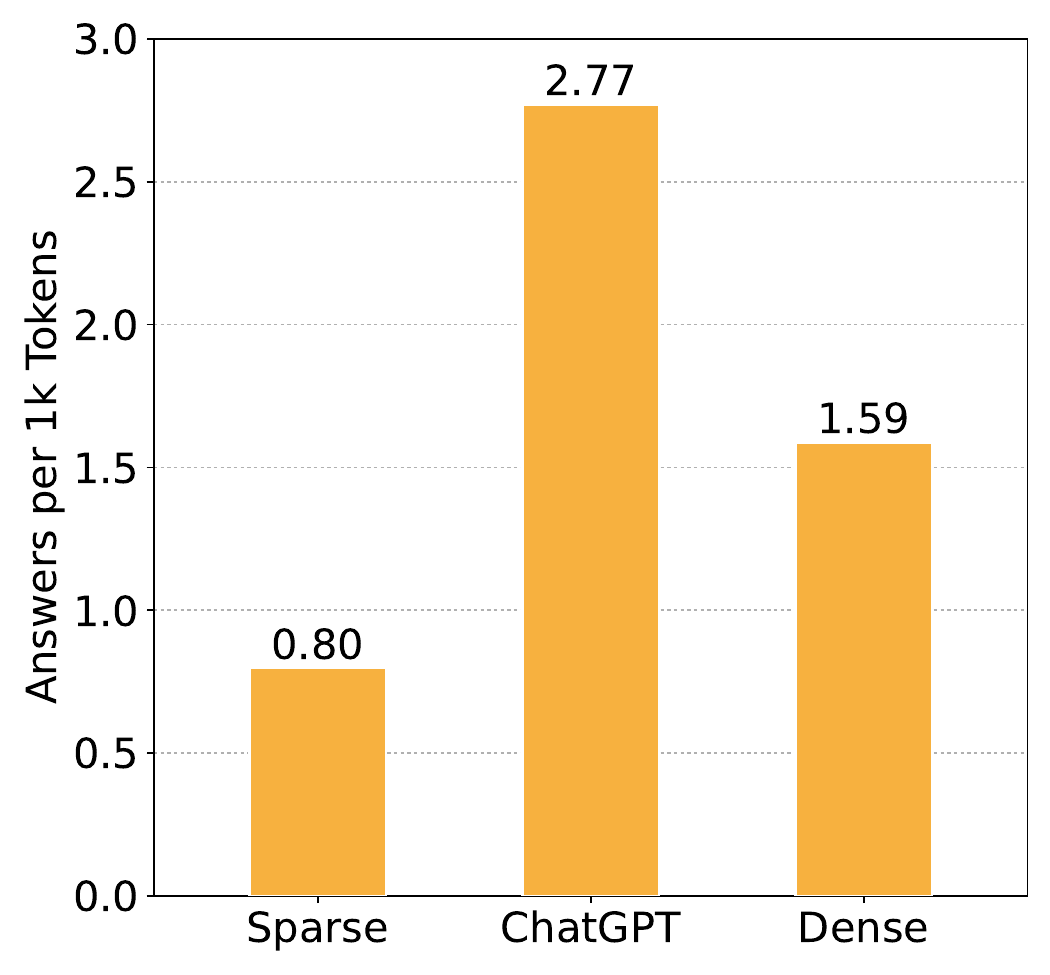}} 
    \subfigure[Exact Match]{\includegraphics[width=0.66\columnwidth]{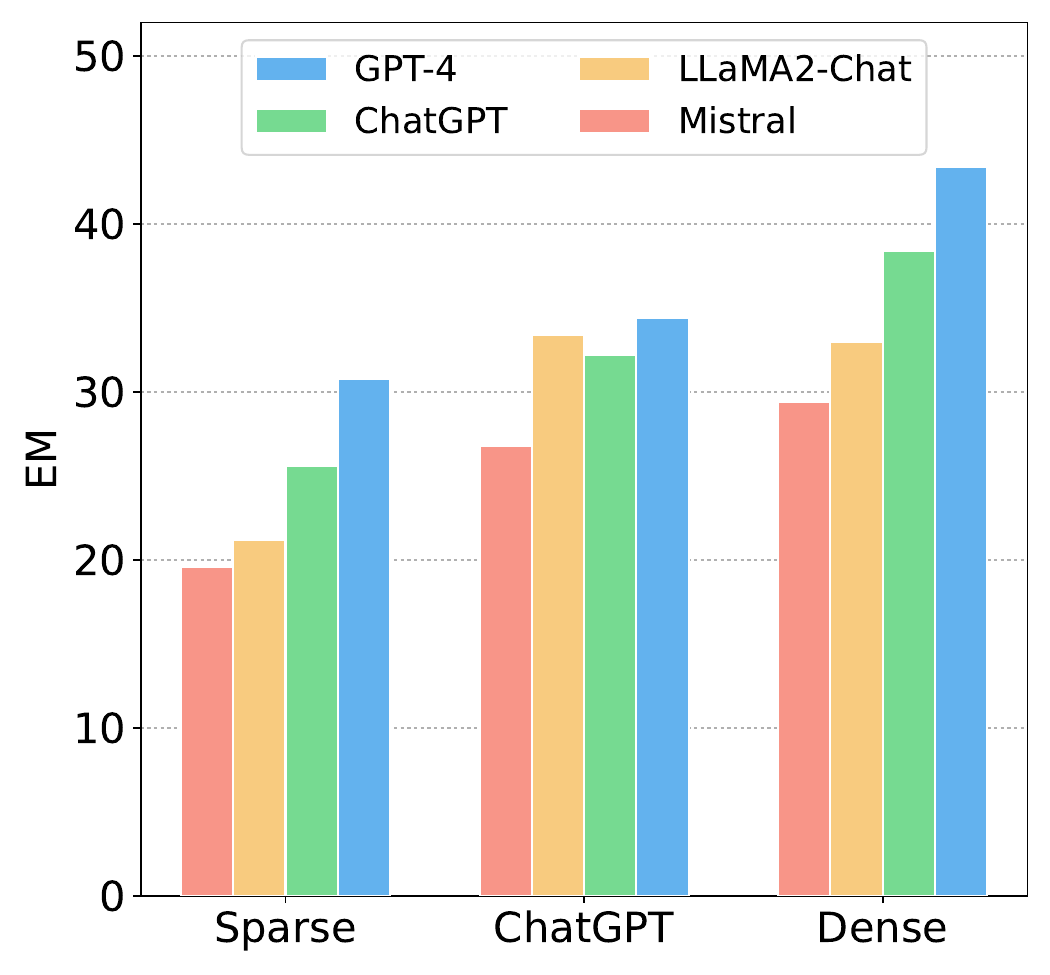}} 
    \subfigure[Give-up Rates]{\includegraphics[width=0.66\columnwidth]{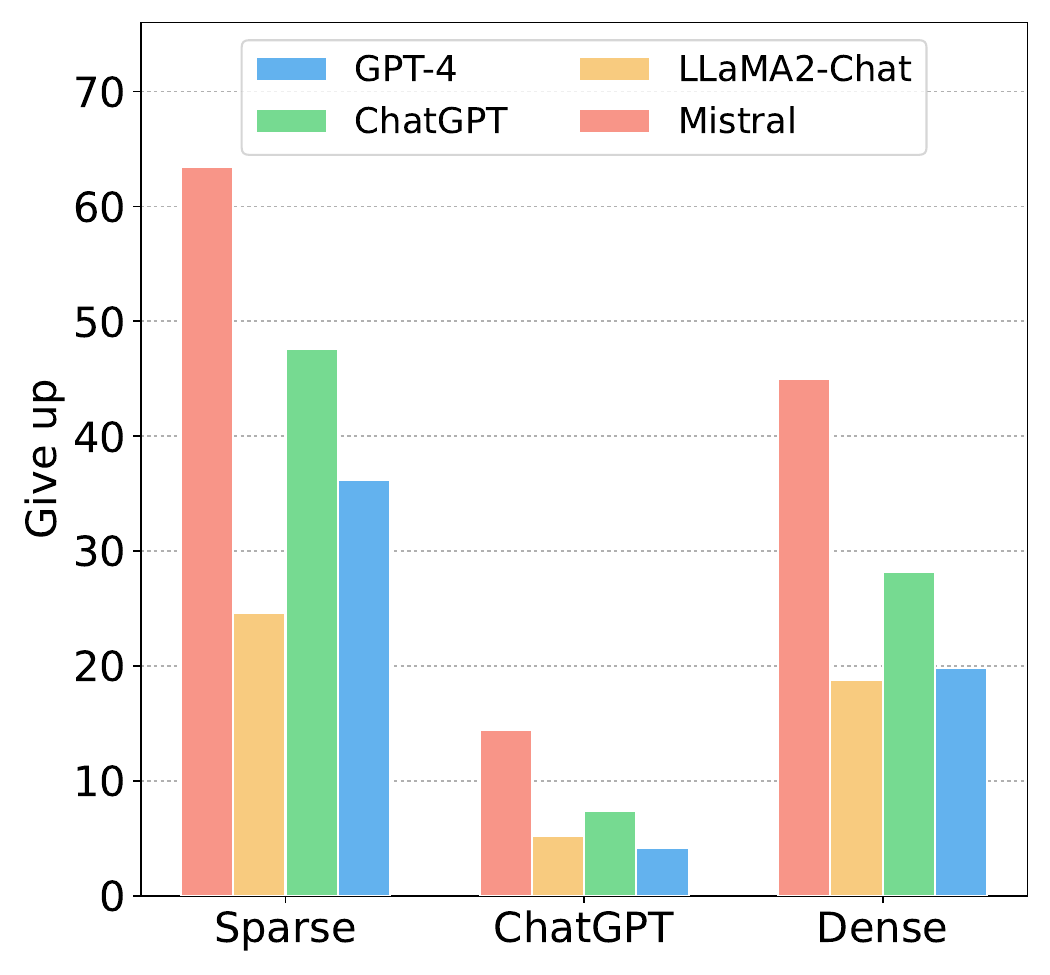}} 
    \subfigure[Eval-Acc]{\includegraphics[width=0.66\columnwidth]{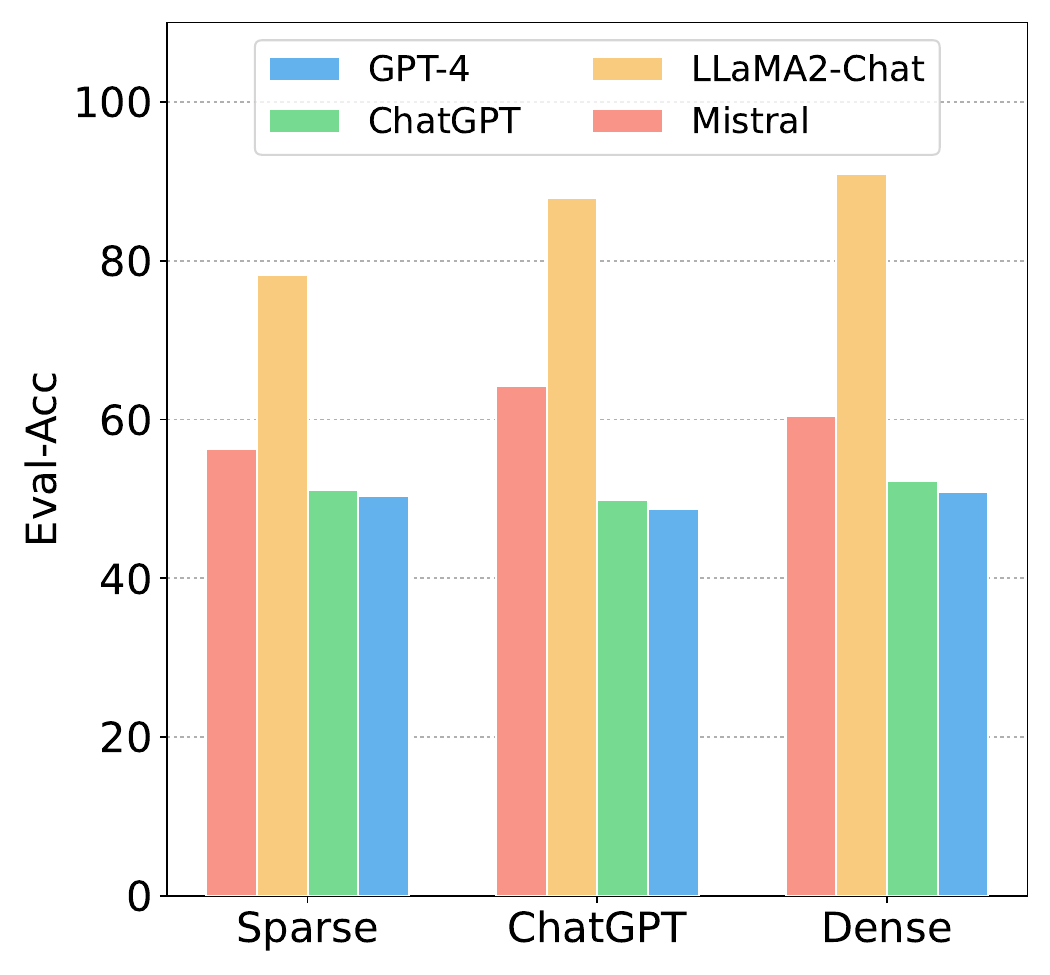}} 
    \caption{Metrics related to retrieval, QA, and judgement using different retrievers.}
    \label{fig:analysis_retriever}
\end{figure*}

\subsection{Analysis on Retrieval Sources}
\label{appendix:analysis_retriever}
\paratitle{Using supporting documents with high-efficiency contents can effectively improve the QA performance of LLMs.}
We conduct the analysis on three retrieval sources and utilize various metrics related to retrieval and QA in Figure~\ref{fig:analysis_retriever} to thoroughly analyze the impact of retrieval augmentation from multiple perspectives. 
Here, the efficiency of documents can be evaluated through three factors: the presence of positive content (recall rate), proportion of positive documents in supporting documents (positive percentage), and token-level text density of positive content (answers per 1K token).
Firstly, the recall rate of documents generated by ChatGPT was significantly lower than that of the dense retriever. However, the QA performance using supporting documents generated by ChatGPT as supporting documents do not lag far behind dense retrieval. Additionally, LLaMA2's retrieval-augmented QA performance with ChatGPT as the retriever exceeds that where the dense retriever is implemented. We observe that among the documents from the three retrieval sources, dense retrieval results have the highest average proportion of positive documents (containing the correct answers), but the number of answers per token was not as high as in documents generated by ChatGPT. This phenomenon stems from the fact that documents generated by ChatGPT are more concise and the documents are closely related to the given questions.
Although Wikipedia documents from the dense retriever have a higher recall rate and proportion of positive documents, they are relatively longer and contain more irrelevant information. As discussed earlier, this could be one of the reasons limiting the upper bound of retrieval augmentation. 
Furthermore, it could potentially contribute to the heightened confidence levels observed in LLMs when utilizing ChatGPT-generated documents as supporting documents, as evidenced by their low \textit{Give-up} rates. In addition, the impact of different retrieval sources on \textit{Eval-Acc} is not substantial.

\subsection{Analysis on Query Types}
\label{appendix:query_type}

\begin{figure*}
    \centering
    \includegraphics[width=0.9\textwidth]{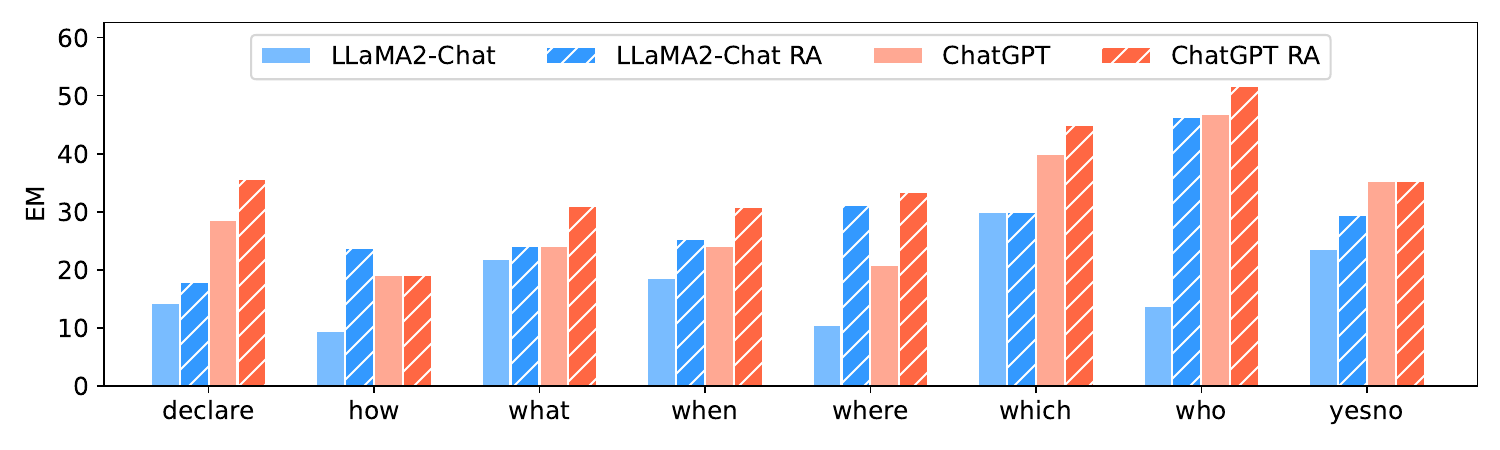}
    \caption{The proportion of questions answered correctly by LLMs in different question categories under two QA prompting settings on NQ.}
    \label{fig:query_type_em}
\end{figure*}

\begin{figure*}
    \centering
    \includegraphics[width=0.9\textwidth]{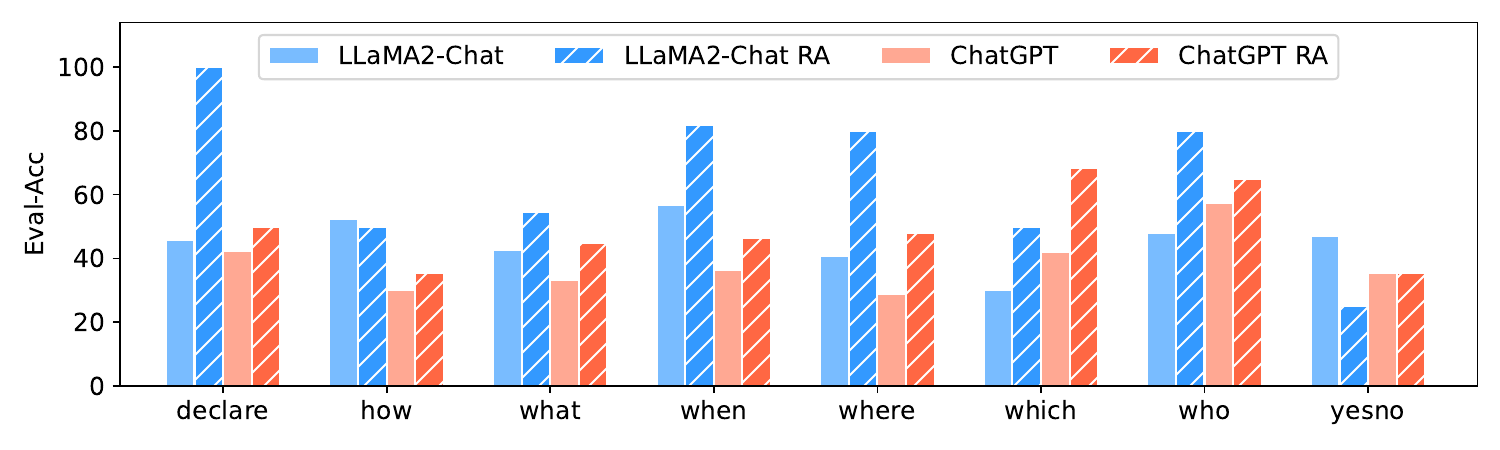}
    \caption{The proportion of answers judged correctly by LLMs in different question categories under two QA prompting settings on NQ.}
    \label{fig:query_type_eval_acc}
\end{figure*}

\paratitle{Retrieval augmentation can change the query category preference of LLMs in QA and factual knowledge boundary perception.}
In order to investigate the propensity of LLMs to handle questions with varied characteristics, we separately calculate the answer accuracy of LLMs across different question categories. To achieve this, we utilize supporting documents retrieved by the dense retriever. As shown in Figure~\ref{fig:query_type_em}, we can see that ChatGPT and LLaMA2 achieve the highest \textit{EM} when dealing with questions in the ``\textit{who}'' and ``\textit{which}'' category, indicating these types of questions may be the strong suit of the two LLMs.
On the other hand, ChatGPT and LLaMA2 may not suffice for the question type of ``\textit{how}'' in knowledge-intensive scenarios.
{When retrieval augmentation is incorporated, we observe that the preference of LLMs changes. 
The overall answer accuracies of LLMs are improved}, and the accuracies in most categories increase proportionately. {Specially, both ChatGPT and LLaMA2 perform best on the question type ``\textit{who}''}. 
We found that \textit{EM} for certain question types does not increase with the overall EM increase (\eg ``how'' and ``yesno'' for ChatGPT, "which" for LLaMA2), indicating that retrieval augmentation cannot effectively enhance the QA performance for all question types.
Figure~\ref{fig:query_type_eval_acc} shows the Eval-Acc metric in each query type for ChatGPT and LLaMA2. We can see that most types of questions achieve better posteriori judgement accuracies, except ``\textit{how}'' and ``\textit{yesno}'' for LLaMA2 and ``\textit{yesno}'' for ChatGPT. Similarly, the result indicates that retrieval augmentation does not uniformly improve the knowledge boundary perception abilities of LLMs across all question types.

\subsection{Analysis on Parameter Scale}
\label{appendix:parameter_scale}

\begin{figure*}
    \centering
    \includegraphics[width=0.98\textwidth]{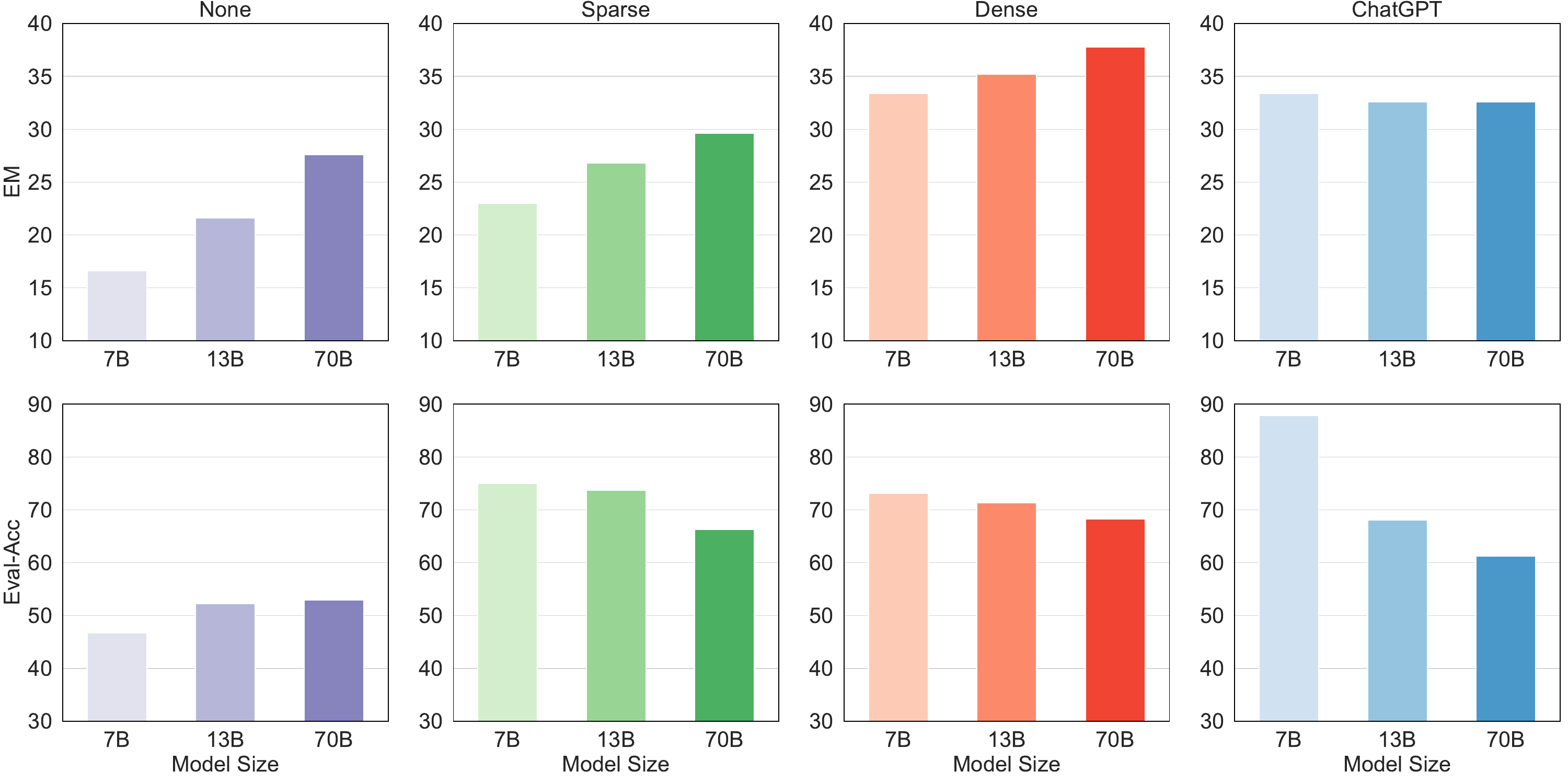}
    \caption{
    The performance of the LLaMA2 series, across its 7B, 13B, and 70B versions, showcases the influence of various retrieval document sources on \textit{EM} and \textit{Eval-Acc}.}
    \label{fig:scale}
\end{figure*}

\paratitle{Retrieval augmentation is more pronounced for improving LLMs with fewer parameters, including both QA performances and accuracies of knowledge boundary perception.}
Figure~\ref{fig:scale} illustrates the comparison of \textit{EM} and \textit{Eval-Acc} before and after retrieval augmentation for LLaMA2 models of varying parameter scales.
We can observe that retrieval augmentation reduces the performance gaps in QA performance (shown by \textit{EM}) of LLMs with different parameter scales, LLaMA2-7B model has the biggest improvement of \textit{EM}. 
It is evident that retrieval augmentation mitigates the performance gaps in QA performance, as demonstrated by \textit{EM} across LLMs with varying parameter scales. Notably, the LLaMA2-7B exhibits the most substantial improvement in \textit{EM} than 13B and 70B models.
Furthermore, the improvement in the accuracy of knowledge boundary perception of the LLaMA2-7B model with retrieval augmentation surpasses that of the 13B and 70B models, achieving the highest \textit{Eval-Acc}.
This phenomenon further supports our speculation that in knowledge-intensive tasks, LLMs with smaller parameter scales possess less knowledge. Under a retrieval augmentation setting, these LLMs are likely to yield greater benefits compared to LLMs with larger parameter scales.

\section{Related Work}
\label{sec:related}
In this section, we review the related work from two perspectives, including large language models~(LLMs) and open-domain question answering~(QA).

\subsection{Large Language Models}

LLMs have significantly advanced the field of natural language processing (NLP), demonstrating remarkable abilities in generating human-like text and understanding complex language patterns~\citep{zhao2023survey}. These models are trained on vast corpora of text data and can perform a wide range of tasks without task-specific training~\cite{thirunavukarasu2023large, lin2024bridging, ren-etal-2024-bases, ren2024self}. 

Typically, LLMs can fall into two categories: publicly available and closed source. 
Publicly available LLMs offer the research community the flexibility to modify and adapt models to specific needs, fostering further innovation and collaboration.
Among publicly available LLMs, the LLaMA series models stands out for its exceptional instruction-following performance~\citep{Touvron-arxiv-2023-LLaMA, Touvron-2023-llama2-arxiv}, with ongoing efforts to fine-tune or continually pre-train its versions.
Mistral, another open-source framework~\citep{jiang-2023-arxiv-mistral}, provides a scalable solution for training LLMs on diverse datasets and supports various architectures to achieve top performance.
Additionally, there are also LLMs like Falcon~\citep{TII-Web-2023-Falcon}, Vicuna~\citep{vicuna2023}, and GLM~\citep{Zeng-arxiv-2022-GLM} actively contribute to advancing publicly available LLMs.	
Conversely, closed source LLMs, developed and maintained by private organizations, have non-public checkpoints and confidential internal workings.
The typical closed source LLMs are GPT series proposed by OpenAI, including initial GPT to the latest GPT-4~\citep{radford-openai-2018-improving, Brown-NeurIPS-2020-Language, OpenAI-OpenAI-2023-GPT-4}, have set new benchmarks in language understanding and generation, finding widespread use in commercial applications. 
There are also other closed source LLMs have been released, such as Sparrow~\citep{Glaese-arxiv-2022-Improving} and Claude. Some closed source LLMs are accessible via API calls, eliminating the need for local execution.

Despite the impressive advancements in LLMs, there remain unexplored boundaries regarding their knowledge and capabilities. Furthermore, the incorporation of retrieval augmentation in LLMs presents substantial potential for enhancing their knowledge, potentially reshaping the knowledge boundaries of LLMs. Therefore, it is necessary to comprehensively investigate the knowledge boundaries of LLMs and the effects of retrieval augmentation.

\subsection{Open-domain Question Answering}

Open-domain QA is a typical knowledge-intensive task in the field of NLP, aimed at providing precise and pertinent answers to questions posed in natural language, without the limitations of a specific domain or topic~\citep{prager2007open}. This task is inherently challenging due to the vastness of human knowledge and the complexity of natural language understanding. Early approaches to open-domain QA rely heavily on structured knowledge bases, such as Freebase~\citep{bollacker2008freebase} or DBpedia~\citep{auer2007dbpedia}, enabling precise yet limited retrieval of factual information. However, these approaches are constrained by the need for structured queries and the coverage of the underlying knowledge bases.
The introduction of pre-trained language models, such as BERT~\citep{bert2019naacl}, BART~\citep{Lewis2020BARTDS}, and their successors, marks a significant shift in the landscape of open-domain QA. These models leverage advanced text modeling techniques to understand natural language, enabling more flexible and context-aware answering performances. 

Recent advancements in open-domain QA have focused on incorporating external knowledge sources, such as Wikipedia or the web, to broaden the scope of answerable questions~\citep{chen17openqa}. This has led to the development of sophisticated information retrieval mechanisms~\citep{ai2023information}, such as dense passage retrieval (DPR)~\citep{dpr2020, rocketqa, zhan2021optimizing, zhuang2021dealing, zhan2022learning} and passage reranking~\citep{nogueira2019passage, multistage, gao2021rethink}, which efficiently extract and prioritize relevant information from large corpora~\citep{DRSurvey, zhu2023large, guo2022semantic}. Additionally, techniques like few-shot learning and transfer learning have been explored to enhance the models' ability to generalize across diverse question types and domains, reducing the reliance on extensive labeled datasets~\citep{brown2020language, REALM, liu2022challenges}.

Despite significant progress, open-domain QA remains an active area of research, with ongoing efforts to improve answer accuracy~\citep{izacard2021leveraging, jin2024bider}, interpretability~\citep{ribeiro2016should}, knowledge utilization~\citep{wang2024rear, zhou2024metacognitive, ni2024llms}, and robustness to complex or ambiguous queries~\citep{lewis2020retrieval, rajpurkar2018know}. Challenges such as handling out-of-domain questions, dealing with evolving knowledge, and ensuring fairness and transparency in answers continue to drive innovation in this field~\citep{gebru2021datasheets, mitchell2019model}. 
Furthermore, Several pioneering studies have been specifically designed to apply LLMs to open-domain QA tasks~\citep{qin2023chatgpt, kamalloo2023evaluating, yue2023automatic, wang2023evaluating, sun2023beamsearchqa, xu2024unsupervised}.
Typically, they mainly focus on evaluating the QA performance of LLMs, discussing improved evaluation methods or leveraging LLMs to enhance existing open-domain QA models. Additionally, the existing effort also detects the uncertainty of LLMs with an automated method~\citep{YinSGWQH23}.
As open-domain QA systems become increasingly integrated into real-world applications, they have potential to revolutionize access to information and knowledge.




\end{document}